\documentclass[a4paper,fleqn]{cas-dc}

\usepackage[numbers]{natbib}

\usepackage{amsmath,amssymb,amsfonts}
\usepackage{setspace}

\graphicspath{{./figs/}}
\DeclareGraphicsExtensions{.pdf,.jpeg,.png}

         \usepackage{amssymb}
\usepackage{bbm} 
\usepackage{algorithmic}
\usepackage{algorithm}
\usepackage{tikz}
\usepackage{comment}
\usepackage{color}
\usepackage{overpic}
\usepackage{graphicx}
\usepackage{pifont}
\usepackage{pict2e} 
\usepackage{booktabs}
\usepackage{multirow}
\usepackage{multicol}
\usepackage{xcolor,colortbl}
\usepackage{nicematrix}

\usepackage{subcaption}

\newcommand{\myPara}[1]{\vspace{.0in}\textbf{#1}.}

\usepackage[capitalize,noabbrev]{cleveref}
\crefname{section}{Sec.}{Secs.}
\crefname{figure}{Fig.}{Figs.}
\crefname{table}{Tab.}{Tabs.}
\crefname{equation}{Eq.}{Eqs.}
\crefname{algorithm}{Alg.}{Algs.}

\definecolor{color1}{RGB}{237,125,49}
\definecolor{color3}{RGB}{0,112,192}
\definecolor{color2}{RGB}{112,173,71}
\usepackage{etoolbox}
\usepackage{listofitems}
\usepackage{wheelchart}
\usetikzlibrary{decorations.text}
\usetikzlibrary{decorations.markings}
\usepackage{makecell}
\usepackage{wrapfig}

\usepackage{forest}
\usetikzlibrary{trees,positioning,shapes,shadows,arrows.meta}

\usetikzlibrary{arrows.meta, shadows, positioning, backgrounds}
\usetikzlibrary{trees}
\definecolor{hidden-draw}{rgb}{0.5, 0.5, 0.5}
\definecolor{harvestgold}{rgb}{0.85, 0.57, 0.0}
\definecolor{cyan}{rgb}{0.0, 1.0, 1.0}
\definecolor{lightcoral}{rgb}{0.94, 0.5, 0.5}
\definecolor{SlateBlue}{rgb}{0.416, 0.353, 0.804}
\definecolor{LightGreen}{rgb}{0.564, 0.933, 0.564}
\definecolor{Turquoise}{rgb}{0.251, 0.878, 0.816}
\definecolor{BlueGreen}{rgb}{0.643, 0.859, 0.871}
\definecolor{BlueViolet}{rgb}{0.54, 0.17, 0.89}
\definecolor{RAG}{rgb}{0.639, 0.835, 0}
\definecolor{0}{rgb}{0.92, 0.3, 0.26}
\definecolor{1}{rgb}{0.961, 0.851, 0.471}
\definecolor{1_1}{rgb}{1.00, 0.933, 0.698}
\definecolor{2}{rgb}{0.522, 0.784, 0.953}
\definecolor{2_1}{rgb}{0.722, 0.592, 0.871}
\definecolor{2_2_2}{rgb}{0.561, 0.659, 0.843}
\definecolor{3}{rgb}{0.56, 0.93, 0.56}
\definecolor{3_1_1}{rgb}{0.808, 0.996, 0.808}
\definecolor{4}{rgb}{0.988, 0.541, 0.565}
\definecolor{4_1}{rgb}{0.996, 0.796, 0.808}
\definecolor{4_1_1}{rgb}{0.957, 0.847, 0.843}
\definecolor{bounding}{rgb}{0.855, 0.392, 0.357}
\definecolor{6}{rgb}{0.690, 0.612, 0.522}

\definecolor{golden}{RGB}{255, 215, 0}
\definecolor{golden}{RGB}{0, 0, 0}

\definecolor{backbone}{RGB}{255,192,0}
\definecolor{neck}{RGB}{112,173,71}
\definecolor{head}{RGB}{237,125,49}
\definecolor{score}{RGB}{68,72,212}
\definecolor{grade}{RGB}{92,201,207}
\definecolor{rank}{RGB}{0,176,207}

\definecolor{OfficeBlue}{RGB}{68,114,196}
\definecolor{OfficeOrange}{RGB}{237,125,49}
\definecolor{OfficeGray}{RGB}{165,165,165}
\definecolor{OfficeYellow}{RGB}{255,192,0}
\definecolor{OfficeLightBlue}{RGB}{91,155,213}
\definecolor{OfficeGreen}{RGB}{112,173,71}

\usepackage{fontawesome}

\usepackage{nicematrix}
\usepackage{calc}

\usepackage{orcidlink}
\usepackage{ragged2e}

\newlength{\myMheight}
\usepackage{bbding,pifont}

\usepackage[misc]{ifsym}

\usepackage{etoolbox}
\makeatletter
\patchcmd{\thebibliography}
  {\settowidth}
  {\setlength{\itemsep}{0pt plus 0.3ex}
   \setlength{\parskip}{0pt}
   \settowidth}
  {}{}
\makeatother

\usepackage[pagewise]{lineno} 

\usepackage{xstring}   

\colorlet{highlightcolor}{red!70!black}
\newif\ifMShighlight
\MShighlighttrue

\newcommand{\MSsubsubsection}[2]{\subsubsection{#2} \label{#1}
}

\makeatletter
\let\MS@originput\input

\renewcommand{\input}[1]{\IfBeginWith{#1}{r3/}{\StrBehind{#1}{r3/}[\MS@strippedfilename]\IfFileExists{#1.tex}{\ifMShighlight \begingroup \color{highlightcolor}\arrayrulecolor{highlightcolor}\linelabel{start:#1}\MS@originput{#1}\linelabel{end:#1}\arrayrulecolor{black}\endgroup
      \else
        \MS@originput{#1}\fi
    }{\MS@originput{\MS@strippedfilename}}}{\MS@originput{#1}}\ignorespaces
}
\makeatother

\def\tsc#1{\csdef{#1}{\textsc{\lowercase{#1}}\xspace}}
\tsc{WGM}
\tsc{QE}
\tsc{EP}
\tsc{PMS}
\tsc{BEC}
\tsc{DE}

\makeatletter
\renewcommand{\vspace}{\@ifstar\@gobble\@gobble}
\makeatother
\usepackage{fancyvrb}
\begin{document}
\let\WriteBookmarks\relax
\def\floatpagepagefraction{1}
\def\textpagefraction{.001}

\shorttitle{A comprehensive survey of action quality assessment: Method and benchmark}

\shortauthors{K. Zhou et~al.}

\title [mode = title]{A comprehensive survey of action quality assessment: Method and benchmark}

\author[1,2]{Kanglei Zhou}
\ead{zhoukanglei@tsinghua.edu.cn}
\credit{Writing -- original draft, Formal analysis}

\author[2]{Ruizhi Cai}
\ead{craaaaazy@buaa.edu.cn}
\credit{Formal analysis, Data curation, Conceptualization}

\author[1]{Liyuan Wang}
\ead{wly19@tsinghua.org.cn}
\credit{Writing -- review \& editing, Conceptualization}

\author[3]{Hubert P.H. Shum}
\ead{hubert.shum@durham.ac.uk}
\credit{Writing -- review \& editing}

\author[2,4]{Xiaohui Liang}[orcid=0000-0001-6351-2538]
\cormark[1]
\ead{liang_xiaohui@buaa.edu.cn}
\credit{Writing -- review \& editing, Supervision}

\cortext[cor1]{Corresponding author}

\affiliation[1]{
    organization={Department of Psychological and Cognitive Sciences, Tsinghua University},
    addressline={No. 30 Shuangqing Road, Haidian District},
    city={Beijing},
    postcode={100084},
    country={China}
}

\affiliation[2]{
    organization={State Key Laboratory of Virtual Reality Technology and Systems, Beihang University},
    addressline={No. 37 Xueyuan Road, Haidian District},
    city={Beijing},
    postcode={100191},
    country={China}
}

\affiliation[3]{
    organization={Department of Computer Science, Durham University},
    addressline={Stockton Rd},
    city={Durham},
    postcode={DH1 3LE},
    country={United Kingdom}
}

\affiliation[4]{
    organization={Zhongguancun Laboratory},
    city={Beijing},
    country={China}
}

\begin{abstract}
Action Quality Assessment (AQA) aims to automatically evaluate how well human actions are performed and has been widely applied in sports analysis, skill assessment, and healthcare. However, AQA studies are often developed under heterogeneous datasets and evaluation settings, making systematic comparison across methods difficult.
To address these challenges, we present a comprehensive survey of recent advances in AQA. In particular, we propose a modality-driven hierarchical taxonomy that organizes existing methods into video-based, skeleton-based, and multi-modal approaches, and analyze the methodological evolution of representative models. We further establish a unified benchmark for representative video-based AQA methods by integrating diverse datasets and standardized evaluation protocols, enabling consistent comparison in terms of both accuracy and computational efficiency. Finally, we analyze emerging research trends, identify key challenges in current AQA research, and outline future directions ranging from near-term methodological advances to longer-term opportunities enabled by emerging AI paradigms. The project webpage is available at \url{https://ZhouKanglei.github.io/AQA-Survey}.
\end{abstract}




\thispagestyle{empty}

\begin{center}
\begin{minipage}{0.94\textwidth}
\Large
\medskip

\noindent\faQuoteLeft\quad
\textbf{Citation and resources.}
This manuscript corresponds to the published review article in
\textit{Pattern Recognition}. Please cite the published version below.

\medskip
\noindent\faLink\quad
\textbf{Published version:}
\href{https://doi.org/10.1016/j.patcog.2026.113933}
{10.1016/j.patcog.2026.113933}

\medskip
\noindent\faGlobe\quad
\textbf{Related resources.}
The project webpage, curated paper list, and benchmark code are available at:

\smallskip
\noindent
\faHome\quad
\href{https://zhoukanglei.github.io/AQA-Survey/}
{Project page}
\qquad
\faGithub\quad
\href{https://github.com/ZhouKanglei/Awesome-AQA}
{Awesome-AQA}
\qquad
\faGithub\quad
\href{https://github.com/ZhouKanglei/AQA-Benchmark}
{AQA-Benchmark}

\medskip
\noindent\faCode\quad
\textbf{BibTeX:}
\end{minipage}
\end{center}

\begin{quote}
\Large
\begin{verbatim}
@article{zhou2026comprehensive,
  author  = {Kanglei Zhou and Ruizhi Cai and Liyuan Wang and
             Hubert P. H. Shum and Xiaohui Liang},
  title   = {A comprehensive survey of action quality assessment:
             Method and benchmark},
  journal = {Pattern Recognition},
  year    = {2026},
  pages   = {113933},
  doi     = {10.1016/j.patcog.2026.113933},
  note    = {Review article}
}
\end{verbatim}
\end{quote}

\begin{center}
\begin{minipage}{0.94\textwidth}
\hrule
\end{minipage}
\end{center}

\medskip

\maketitle

\section{Introduction} 
\label{sec:intro}
Action Quality Assessment (AQA)~\cite{liu2025aqasurvey,yin2025decade} aims to automatically assess how well an action is performed. Unlike action recognition, which focuses on identifying different action categories, AQA evaluates performance differences within the same action category. Consequently, accurately assessing action quality requires sensitivity to subtle variations in motion dynamics and domain-specific evaluation criteria, making it significantly more challenging than traditional recognition tasks. By providing an objective alternative to subjective judgments, AQA has been widely applied in domains such as sports analysis~\cite{xu2019learning,parmar2017learning}, skill assessment~\cite{parmar2021piano,zhang2014relative}, and healthcare~\cite{capecci2019kimore,zhou2023video}. More broadly, AQA can support intelligent perception systems by enabling quantitative evaluation of action performance, facilitating applications such as skill coaching and human--robot collaboration in embodied AI environments~\cite{huang2024egoexolearn,wu2025skillsight}. 
Since AQA spans different domains with varying terminology, this survey uses \emph{AQA} as the primary term, encompassing related concepts such as action scoring, skill assessment, and performance evaluation. 
\begin{figure}[!ht]
    \centering
    \begin{minipage}[c]{\linewidth}
        \centering
        \includegraphics[width=\linewidth]{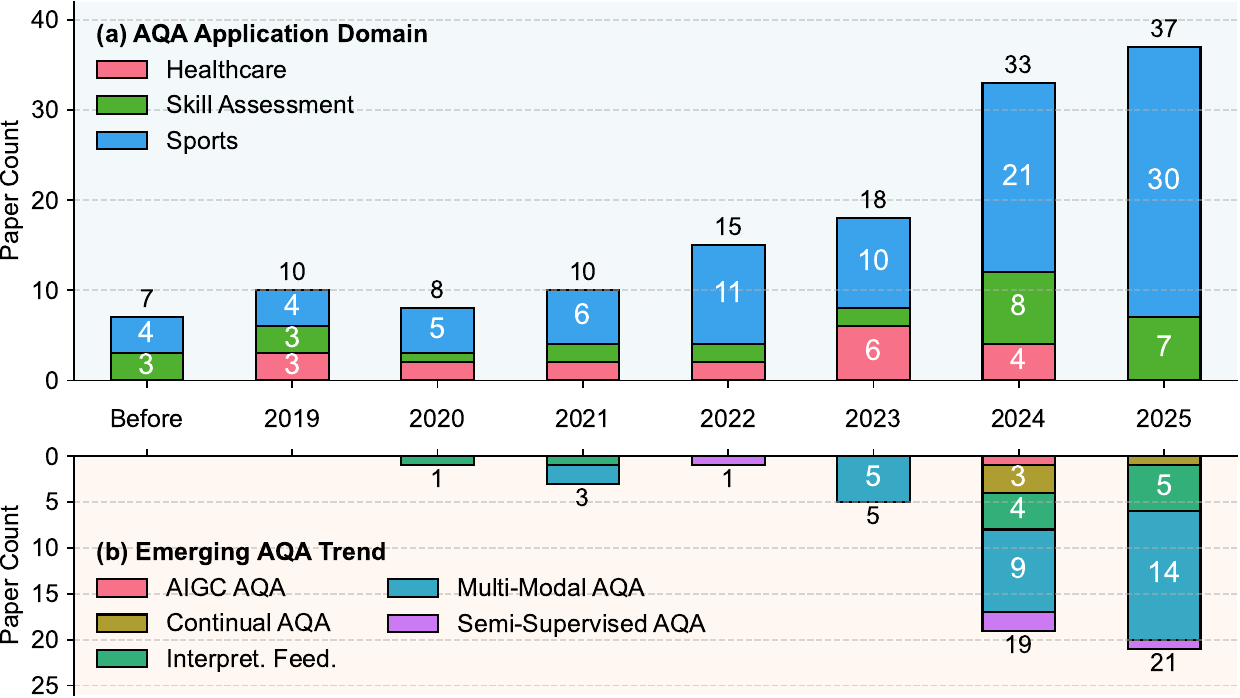}
    \end{minipage}
    \hfill 
    \begin{minipage}[c]{\linewidth}
        \caption{Annual statistics of methodological AQA papers in major CV and ML venues included in this survey. Papers are categorized by (a) application domains and (b) research directions, which have not been systematically summarized in prior AQA surveys.}
        \label{fig:paper_count}
        {
        \phantomsubcaption\label{fig:paper_count-a}\phantomsubcaption\label{fig:paper_count-b}}
    \end{minipage}
    \vspace{-0.25cm}
\end{figure} 

Recent advancements in deep learning~\cite{wang2024comprehensive,chang2025design} have driven substantial progress in AQA (see \cref{fig:paper_count-a}), enabling increasingly powerful methodologies and expanding application scenarios (see \cref{fig:paper_count-b}). Although several surveys~\cite{lei2019survey,wang2021survey,liu2025aqasurvey,yin2025decade} have summarized AQA research (see \cref{tab:survey-comparison}), they remain limited in scope and methodological analysis.
Early surveys primarily adopt domain-driven taxonomies~\cite{lei2019survey,wang2021survey}, which fail to capture shared methodological foundations across tasks, while trend-oriented reviews~\cite{yin2025decade} summarize research developments but do not explicitly unify overlapping techniques across paradigms. In addition, none of the existing surveys provides a unified benchmark for systematic comparison of AQA methods. Furthermore, recent surveys~\cite{liu2025aqasurvey,yin2025decade} mainly summarize existing studies but provide limited synthesis of emerging research trends and little discussion of future directions. These limitations reveal a fundamental gap in the current literature: existing surveys mainly summarize AQA studies from domain or trend perspectives and lack a unified methodological framework that organizes modeling paradigms and supports consistent evaluation across methods.

As a result, AQA research remains fragmented across datasets, application domains, and modeling paradigms (see \cref{fig:paper_count}), making it difficult to obtain a systematic understanding of the field. Such fragmentation manifests at multiple levels: heterogeneous datasets and evaluation protocols hinder reproducibility; methods are often validated in narrowly defined settings, limiting generalization; and the diversity of application domains obscures shared modeling principles across tasks. This fragmentation highlights the need for a more systematic and unified perspective on AQA research.

\begin{table*}[!t]
    \centering
    \caption{Comparison of existing AQA surveys.
    \includegraphics[height=\myMheight]{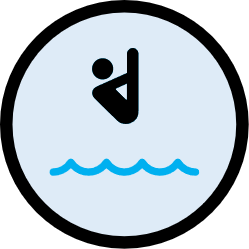}, \includegraphics[height=\myMheight]{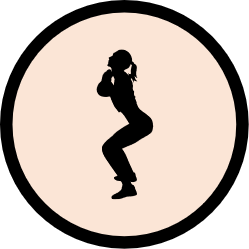}, and \includegraphics[height=\myMheight]{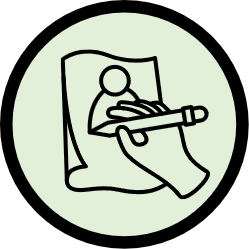} indicate sports, medical care, and skill assessment icons.
    $^{\color{black}\dagger}$ and $^{\color{black}\ddagger}$ denote two-level taxonomy standards.
    \includegraphics[height=\myMheight]{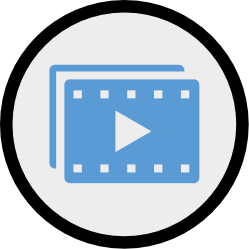}, \includegraphics[height=\myMheight]{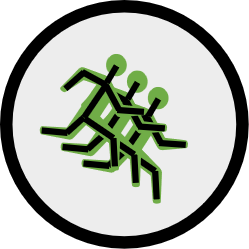}, \includegraphics[height=\myMheight]{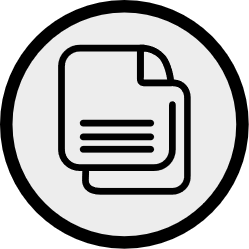}, \includegraphics[height=\myMheight]{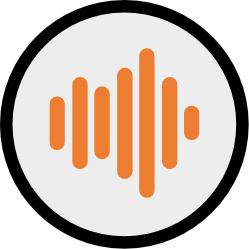}, \includegraphics[height=\myMheight]{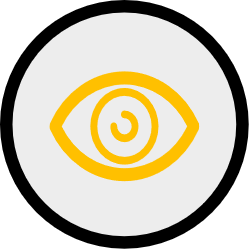}, \includegraphics[height=\myMheight]{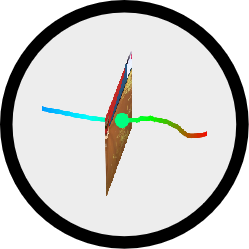}, and \includegraphics[height=\myMheight]{skill-assessment-icon.png} represent video, skeleton, text, audio, gaze, and flow icons.
    \includegraphics[height=\myMheight]{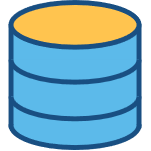} and \includegraphics[height=\myMheight]{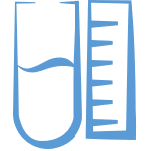} denote the dataset and metric icons.
    }
    \label{tab:survey-comparison}
\rowcolors{2}{gray!3}{gray!12}
    \resizebox{\linewidth}{!}{
    \begin{tabular}{rcm{1.8cm}m{4.5cm}m{2.2cm}m{2.5cm}m{8.5cm}}
    \toprule
    \textbf{Survey}  &  \textbf{Year} & \textbf{Domains} & \textbf{Taxonomy} & \textbf{Modality} & \textbf{Benchmark}  & \textbf{Core Contributions} \\
    \midrule
    Lei et al. \cite{lei2019survey}  & 2019 & \includegraphics[width=0.45cm]{sports-icon.png} \includegraphics[width=0.45cm]{medical-care-icon.png}
    \includegraphics[width=0.45cm]{skill-assessment-icon.png}
    & $^{\color{black}\dagger}$ Feature type, $^{\color{black}\ddagger}$ Domain  & 
    \includegraphics[width=0.45cm]{video-icon.png}
    \includegraphics[width=0.45cm]{skeleton-icon.png}
    & 
None
    & An early overview of motion detection and data preprocessing techniques relevant to AQA. \\
    Wang et al. \cite{wang2021survey}  & 2021 & \includegraphics[width=0.45cm]{sports-icon.png} \includegraphics[width=0.45cm]{medical-care-icon.png} 
    & $^{\color{black}\dagger}$ Domain, $^{\color{black}\ddagger}$ Feature type & 
    \includegraphics[width=0.45cm]{video-icon.png}
    &  
None
    & A domain-oriented survey emphasizing categorization by application fields and publishing venues. \\
    Liu et al. \cite{liu2025aqasurvey}  & 2024 & \includegraphics[width=0.45cm]{sports-icon.png} \includegraphics[width=0.45cm]{medical-care-icon.png}
    \includegraphics[width=0.45cm]{skill-assessment-icon.png} 
    & $^{\color{black}\dagger}$ Input mode & 
    \includegraphics[width=0.45cm]{video-icon.png}
    \includegraphics[width=0.45cm]{skeleton-icon.png}
    &  
None
    & A descriptive review of representative AQA studies, summarizing methods and application scenarios. \\
    Yin et al. \cite{yin2025decade}  & 2025 & \includegraphics[width=0.45cm]{sports-icon.png} \includegraphics[width=0.45cm]{medical-care-icon.png}
    \includegraphics[width=0.45cm]{skill-assessment-icon.png} 
    & $^{\color{black}\dagger}$ Research trends & 
    \includegraphics[width=0.45cm]{video-icon.png}
    \includegraphics[width=0.45cm]{skeleton-icon.png}
    \includegraphics[width=0.45cm]{text-icon.png}
    \includegraphics[width=0.45cm]{audio-icon.png}
    &  
None
    & A trend-oriented survey, highlighting research challenges with a high-level discussion of evaluation practices.  \\
Ours & 2026 & 
    \includegraphics[width=0.45cm]{sports-icon.png} \includegraphics[width=0.45cm]{medical-care-icon.png}
    \includegraphics[width=0.45cm]{skill-assessment-icon.png} 
    & $^{\color{black}\dagger}$ Input mode, $^{\color{black}\ddagger}$ Mode specialty & 
    \includegraphics[width=0.45cm]{video-icon.png}
    \includegraphics[width=0.45cm]{skeleton-icon.png}
    \includegraphics[width=0.45cm]{text-icon.png}
    \includegraphics[width=0.45cm]{audio-icon.png}
    \includegraphics[width=0.45cm]{gaze-icon.png}
    \includegraphics[width=0.45cm]{flow-icon.png}
    \includegraphics[width=0.45cm]{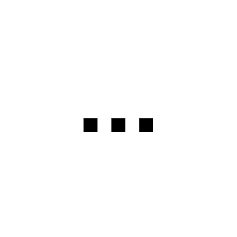}
     & 
6 \includegraphics[height=\myMheight]{dataset.png} with 7 \includegraphics[height=\myMheight]{metric.png}\par
     Unified setting
     & A modality-driven hierarchical taxonomy, a unified benchmark, specific AQA tasks, challenges, and prospects. \\
    \bottomrule
    \end{tabular}
    }
\end{table*}
  
To bridge this gap, we present a systematic review of recent advances in AQA.
\textbf{First}, we argue that the type of input data, rather than the application domain, is the primary determinant of AQA model design. 
Based on this premise, we propose a modality-driven hierarchical taxonomy. Specifically, we categorize approaches into video-based, skeleton-based, and multi-modal methods, further distinguishing them by modality-specific attributes.
\textbf{Second}, we establish a unified AQA benchmark, to the best of our knowledge, integrating six widely used datasets and seven evaluation metrics. By standardizing experimental settings, we facilitate consistent comparisons across diverse methods, evaluating both accuracy and the often-overlooked computation overhead.
\textbf{Third}, beyond summarizing existing work, we analyze emerging trends and outline both near-term and longer-term directions for AQA. Near-term directions focus on improving reliability and interpretability, while longer-term opportunities arise from emerging paradigms such as generative models and embodied AI.

This survey targets researchers developing AQA systems in domains such as sports analytics and healthcare, as well as related areas including human–computer interaction, where objective action evaluation supports decision-making. The remainder of this paper is organized as follows. \Cref{sec:review_method} describes the review methodology, \Cref{sec_unified_framework} introduces the foundational framework, \Cref{sec:taxonomy} analyzes representative AQA methods, \Cref{sec:app} reviews task-specific applications, \Cref{sec:db} summarizes representative datasets and presents the benchmark, \Cref{sec:discussion} discusses emerging trends, remaining challenges, and future research directions, and \Cref{sec:conclusion} concludes the whole paper.

\section{Review Methodology}
\label{sec:review_method}
\begin{figure*}[!ht]
    \centering
    \includegraphics[width=\linewidth,clip,trim=10 20 10 10]{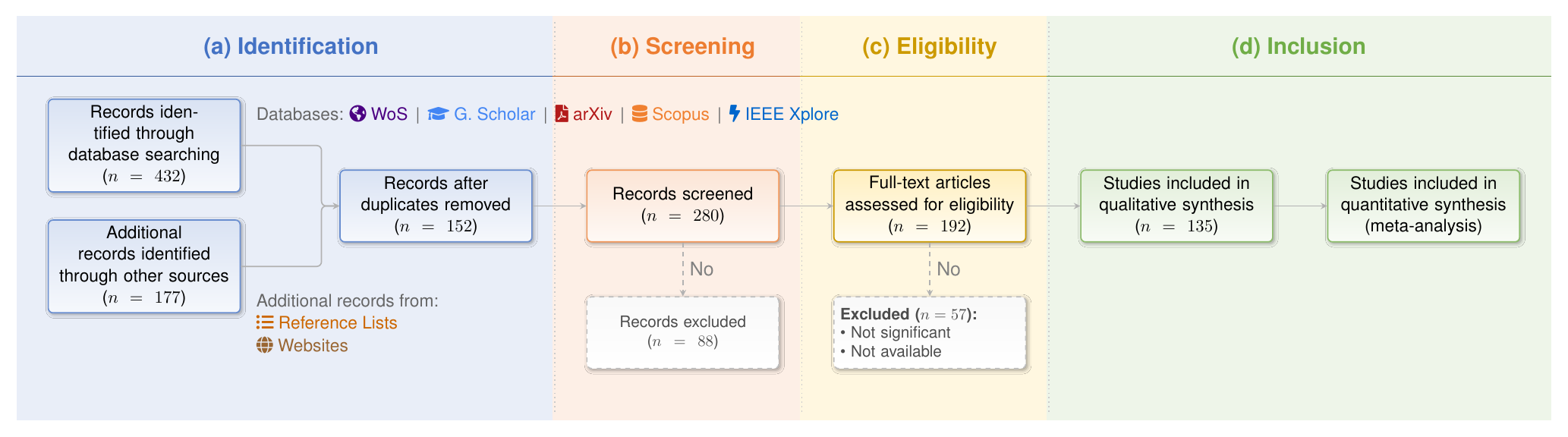}
    \caption{PRISMA flow diagram: (a) identification, (b) screening, (c) eligibility, and (d) inclusion.}
    \label{fig:prisma}
\end{figure*}

To ensure consistent and fair comparison across diverse AQA application domains, we follow the PRISMA framework (see \cref{fig:prisma}) to improve transparency, reproducibility, and reliability in selecting representative studies for this survey.

\vspace{-0.25cm}

\myPara{Identification Strategy}
We searched IEEE Xplore, Web of Science, Scopus, Google Scholar, and arXiv using keywords such as \emph{action quality assessment}, \emph{skill assessment}, and \emph{action scoring}. Publications from 2014 to present were considered, spanning from early handcrafted to recent deep learning-based approaches. We further conducted backward citation analysis on representative papers to improve coverage.

\myPara{Inclusion and Exclusion Criteria}
Studies were included if they: (1) addressed AQA in terms of score, grade, or rank prediction; (2) proposed learning-based methods, benchmark datasets, or evaluation frameworks directly related to AQA; and (3) were peer-reviewed papers published in leading journals or conferences in computer vision and pattern recognition (e.g., TPAMI, IJCV, PR, CVPR, ICCV, ECCV), or widely recognized arXiv preprints with distinct technical contributions.
To ensure quality, we consider four criteria: (1) methodological clarity, (2) experimental rigor, (3) reproducibility, and (4) benchmark relevance. 
Benchmark relevance and methodological clarity are treated as essential inclusion criteria, and studies failing to meet them are excluded. 
Experimental rigor and reproducducibility are used for AQA and prioritization when multiple studies address similar problems. 
We further exclude non-English, tutorial, abstract-only, duplicate studies, and works without explicit quality prediction.%

\myPara{Screening and Selection Process}
After deduplication, titles and abstracts were first screened to remove irrelevant records, followed by full-text assessment based on the above criteria. Papers with insufficient technical details or unclear experimental validation were excluded during full-text review. Disagreements were resolved through discussion among the authors. This resulted in 135 methodological AQA papers included in our final survey (see \cref{fig:prisma}).

\myPara{Data Extraction and Categorization}
From each study, we extracted input modalities, model architectures, supervision types, datasets, and application domains. These attributes support our taxonomy and analysis across AQA methods.

\myPara{Reproducibility and Resources}
To promote transparency, we maintain a public webpage with curated papers and metadata, which will be continuously updated.
 
\section{Fundamentals of Common AQA Setup} \label{sec_unified_framework}
Although AQA spans diverse domains, most methods share a common framework. To enable comprehensive analysis, this section formulates a unified framework for multi-modal inputs and introduces typical scenarios and evaluation metrics.

\begin{figure}[!ht]
    \centering
    \begin{minipage}[c]{\linewidth}
        \centering
        \includegraphics[width=\linewidth,clip,trim=0 58 0 90]{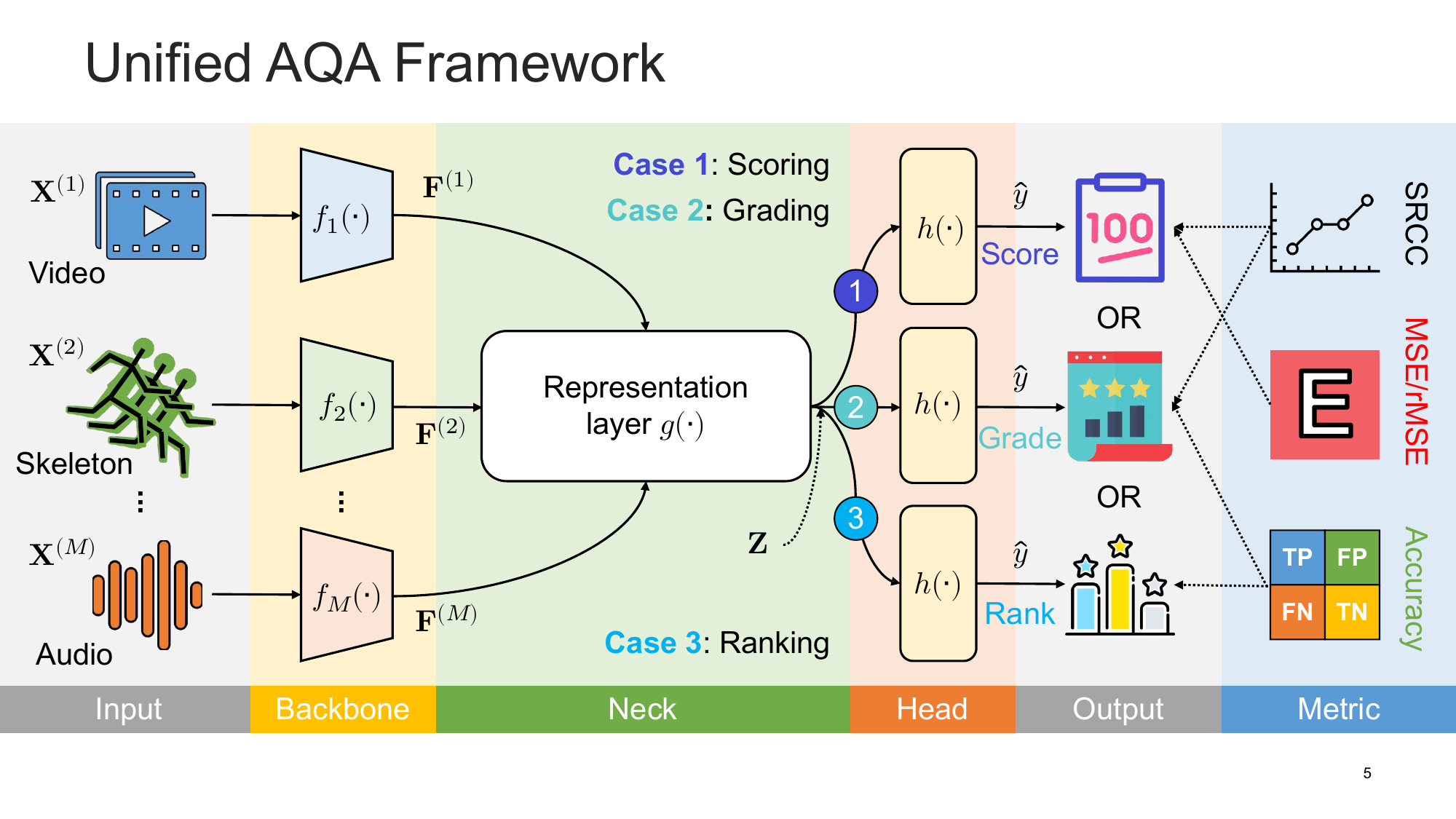}
    \end{minipage}
    \hfill 
    \begin{minipage}[c]{\linewidth} 
        \caption{Three-stage AQA framework: (1) a {\color{backbone}backbone} for modality-aware feature extraction; (2) a {\color{neck}neck} for representation learning that derives compact embeddings; and (3) a regression or classification {\color{head}head} that outputs continuous scores ({\color{score}Case~1}), discrete grades ({\color{grade}Case~2}), or ranks ({\color{rank}Case~3}).}
        \label{fig:unified-aqa-framework}
    \end{minipage}
    \vspace{-0.5cm}
\end{figure}

\subsection{Problem Formulation} \label{sec_unified_framework-aqa_formulation}
\cref{fig:unified-aqa-framework} illustrates the unified framework for AQA, focusing on evaluating the quality of actions through diverse input modalities, including video, skeleton data, and sensor inputs. 
Let
$
\mathcal{X}=\left\{\mathbf{X}^{(1)}, \mathbf{X}^{(2)}, \cdots, \mathbf{X}^{(M)}\right\}
$
represent a set of $M \geq 1$ input modalities, where each
$
\mathbf{X}^{(m)}=\left[\mathbf{x}_1^{(m)}; \mathbf{x}_2^{(m)}; \cdots; \mathbf{x}_T^{(m)}\right]
$
corresponds to a sequence of observations for modality $m$ over time $t \in \{0,1, \cdots, T\}$.
The AQA process in a neural network architecture can be divided into three primary components: feature extraction (backbone), representation learning layer (neck), and score prediction (head).
Each input modality $\mathbf{X}^{(m)}$ is processed through a feature extractor $f_m(\cdot)$, generating modality-specific features
$
\mathbf{F}^{(m)}=f_m\left(\mathbf{X}^{(m)}\right)
$.
Then the extracted features are passed through the neck $g(\cdot)$, which fuses and transforms the features into an overall representation
$
    \mathbf{Z}=g\left(\mathbf{F}^{(1)}, \mathbf{F}^{(2)}, \cdots, \mathbf{F}^{(M)}\right)
$.
Finally, the learned representation $\mathbf{Z}$ is passed through a quality prediction head $h(\cdot)$ to estimate the predicted quality score 
\begin{equation} \label{eq:def}
\hat{y}
=h\left(g\left(f_1\left(\mathbf{X}^{(1)}\right), f_2\left(\mathbf{X}^{(2)}\right), \cdots, f_M\left(\mathbf{X}^{(M)}\right)\right)\right).
\end{equation}
Our formulation retains generality and can handle both unimodal and multi-modal AQA methods. For unimodal methods ($M=1$), the focus is on a single input type, such as video (see \cref{sec:taxonomy-video_based}) or skeletal data (see \cref{sec:taxonomy-skeleton_based}), each of which provides full human-centric cues crucial for AQA. For multi-modal methods ($M>1$), auxiliary modalities aid in comprehensive analysis (see \cref{sec:taxonomy-multi_modality}).

\subsection{Typical Scenarios} \label{sec_unified_framework-aqa_scenarios}
In AQA, the form of the output varies depending on the requirements of different applications. Broadly, AQA methods typically produce continuous or discrete outputs.

\begin{table*}
    \rowcolors{2}{gray!3}{gray!12}
    \centering
    \caption{
    Multidimensional comparison of different AQA scenarios.
    }
    \resizebox{0.85\linewidth}{!}{
    \begin{tabular}{m{1.4cm}m{0.5cm}m{2.0cm}m{6cm}m{6cm}m{6cm}}
    \toprule
    \textbf{Scenarios} & \multicolumn{2}{c}{\textbf{Outputs}} & \textbf{Precision of Feedback} & \textbf{Complexity of Annotation}  & \textbf{Application Suitability}  \\
    \midrule
    Continuous Scoring & 
    \includegraphics[width=0.45cm]{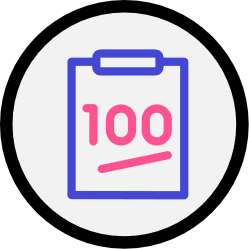}
    & 
    $\hat{y} \in [a, b]$
    &
    \faStar\faStar\faStar: Delivers highly exact feedback under certain criteria
    &
    \faStar\faStar\faStar: Requires precise annotations often from domain experts
    & 
    Sports (e.g., gymnastics \cite{xu2022likert}, diving \cite{parmar2019and}, figure skating \cite{liu2023figure})
    \\
Discrete Grading  & 
    \includegraphics[width=0.45cm]{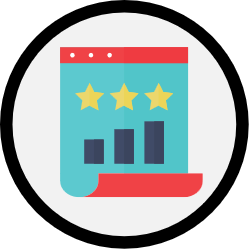} &
    $\hat{y} \in \{1,2,\cdots\}$
    &
    \faStar\faStar\faStarO: Offers output that is easier to interpret but less granular
    &
    \faStar\faStar\faStarO: Demands annotations with moderate granularity
    & Skill assessment (e.g., surgery \cite{wang2020towards}, physical exercise \cite{capecci2019kimore})\\
    Discrete Ranking & 
    \includegraphics[width=0.45cm]{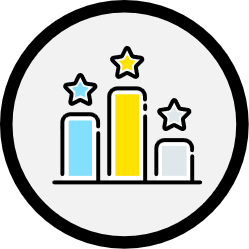} &
    $\hat{y} \in \{ -1, 0, 1 \}$  & 
    \faStar\faStarO\faStarO: Provides general feedback with limited detail
    &
    \faStar\faStarO\faStarO: Requires minimal effort
    & Comparative tasks (e.g., cooking \cite{doughty2018s}, basketball \cite{bertasius2017baller}, teaching \cite{fang2024better}) \\
    \bottomrule
    \end{tabular}
    }
    \label{tab:aqa_scenarios}
\end{table*}

\subsubsection{Continuous Score} 
Continuous scores are commonly used in scenarios that require precise and detailed evaluation, such as in Olympic sports AQA \cite{liu2023figure,parmar2019and,xu2022likert,parmar2017learning,xu2019learning}. The score $\hat{y}$ in \cref{eq:def} typically ranges within a specific interval, $[a,b]$ (e.g., $[0, 100]$), reflecting the exact quality of the action. The primary objective is to predict a continuous numerical value that accurately quantifies action quality, providing precise feedback essential for competitive evaluation.

\subsubsection{Discrete Grade or Rank}
Discrete outputs, such as grade and rank, simplify AQA by mapping performances to levels or relative orders, which is helpful when precise scores are unnecessary.
\textbf{Grading} assigns each performance to a discrete level (e.g., ``excellent", ``good", ``fair", ``poor")~\cite{ding2023sedskill,li2018scoringnet}, providing interpretable and coarse evaluation suited for applications like skill assessment.
\textbf{Ranking} compares performances to establish a relative order, useful for pairwise or group comparisons~\cite{doughty2018s,fang2024better,luo2024rhythmer} where absolute scores are less critical. Given paired samples $(i, j)$, the model outputs:
\begin{equation} \label{eq:ranking}
    \hat{y}_{(i,j)} = 
    \begin{cases}
        1, & i ~\text{outperforms}~ j\\
        -1, & i ~\text{underperforms}~ j\\
        0, & \text{no preference}
    \end{cases}.
\end{equation}

\subsubsection{Implications for Modeling}
As summarized in \cref{tab:aqa_scenarios}, different AQA scenarios impose different supervision and modeling needs. 
\emph{Continuous scoring} treats AQA as fine-grained regression for high-stakes settings but requires precise and robust labels; 
\emph{discrete grading} favors efficiency and interpretability when coarse feedback suffices; 
\emph{ranking} relies on relative comparisons, reducing annotation cost and emphasizing ordinal consistency. 
These choices reflect trade-offs between precision, cost, and robustness, motivating the joint use of correlation- and precision-based metrics in \cref{sec_unified_framework-evaluation_metrics}.

\subsection{Evaluation Metrics} 
\label{sec_unified_framework-evaluation_metrics}
We mainly discuss metrics for evaluating prediction performance in AQA, while computational efficiency metrics are considered later in the unified benchmark.

\subsubsection{Correlation Metrics}
The correlation metric in AQA is the Spearman Rank Correlation Coefficient (SRCC). This metric assesses the strength and direction of the relationship between the predicted scores $\hat{y}_i~(i\in \{1,2,\cdots,N\})$ and the true scores $y_i$ based on their ranks $\hat{r}_i$ and $r_i$. SRCC can be defined as:
\begin{equation} \label{eq:srcc}
\text{SRCC} = \frac{ \sum_{i=1}^{N} \left( r_i - \bar{r} \right) \left( \hat{r}_i - \bar{\hat{r}} \right) }{ \sqrt{\sum_{i=1}^{N} \left( r_i - \bar{r} \right)^2 } \sqrt{\sum_{i=1}^{N} \left( \hat{r}_i - \bar{\hat{r}} \right)^2 } },
\end{equation}
where $\bar{r}$ and $\bar{\hat{r}}$ are the mean ranks of true and predicted scores, respectively.
SRCC is particularly useful in scenarios where the precise numerical score is less important than the relative ranking of actions. This makes it a robust metric for evaluating ordinal data, as it emphasizes the consistency of rankings.

\subsubsection{Precision Metrics} 
Two common types of precision metrics assess the accuracy of predicted scores, tailored to either continuous or discrete outputs (see \cref{sec_unified_framework-aqa_scenarios}). 

\textbf{Score error} quantifies the difference between the predicted and actual scores, allowing for an assessment of the model's accuracy in predicting continuous quality scores. Common measures include Mean Absolute Error (MAE) and Mean Squared Error (MSE), which provide insights into the magnitude of prediction errors. Recently, relative Mean Squared Error (rMSE) \cite{yu2021group,zhou2023hierarchical} has gained popularity as it avoids the impact of differing score scales across actions in different categories. rMSE is:
\begin{align} \label{eq:rmse}
\text{rMSE} = \frac{1}{N} \sum_{i=1}^{N} \left( \frac{y_i - \hat{y}_i}{y_{\max} - y_{\min}} \right)^2 \times 100,
\end{align}
where $y_{\max}$ and $y_{\min}$ denote the maximum and minimum scores, respectively. Note that our definition differs slightly from the relative $\ell_2$ in \cite{yu2021group}. Normalization scales the score to a decimal, and squaring reduces it further. To address this, we multiply by 100 to rescale it for a precise comparison.

\textbf{Accuracy} measures the proportion of correctly predicted grades or ranked pairs relative to the total predictions made for applications where actions are categorized into discrete grades or ranks, which is: 
\begin{equation} \label{eq:acc}
\text{Accuracy} = \frac{ \sum_{i=1}^{N} \mathbbm{1} \left( \hat{y}_i = y_i \right) }{N} \times 100\%,
\end{equation}
where $\mathbbm{1}(\cdot)$ denotes the indicator function, which equals 1 if $\hat{y}_i = y_i$ and 0 otherwise. 

\MSsubsubsection{sec:task-specific-metrics}{Task-Specific Metrics}
Beyond general AQA, task-specific metrics are often required to reflect application objectives.
For AQA with temporal segmentation \cite{xu2022finediving,xu2024procedure,xu2024fineparser}, Intersection over Union (IoU) evaluates segment-level localization quality, which is crucial for fine-grained feedback but less indicative of overall scoring accuracy. For continual AQA \cite{zhou2024magr,zhou2025continual}, average forgetting measures knowledge retention over time, highlighting the trade-off between stability and adaptability to new actions. 

\MSsubsubsection{sec_unified_framework-tradeoff}{Metric Trade-offs across Scenarios}
Different metrics capture complementary aspects and are suited to different scenarios. For ranking-based AQA, correlation metrics (e.g., SRCC) are preferred as they emphasize relative ordering but may tolerate large absolute errors. 
For scoring or grading with continuous or discrete outputs, precision metrics (e.g., rMSE, accuracy) are essential to measure numerical or categorical correctness, though they may not fully capture ranking consistency. 
Task-specific metrics further tailor evaluation to specialized goals but reduce cross-task comparability. 
Thus, many AQA settings jointly adopt correlation- and precision-based metrics. 

\section{A Modality-Driven Taxonomy of AQA Methods}
\label{sec:taxonomy}
A clear taxonomy is essential to address fragmentation in AQA methods. We organize AQA approaches by input modality rather than domains, as it directly determines network design and modeling strategy (see \cref{sec:taxonomy-overview}). Accordingly, methods are grouped into video-based, skeleton-based, and multi-modal categories (see \cref{sec:taxonomy-video_based,sec:taxonomy-skeleton_based,sec:taxonomy-multi_modality}), and we discuss how modality-specific trade-offs between data acquisition cost, representation capability, and robustness guide AQA system design (see \cref{sec:taxonomy-guidline}). 

\subsection{Modality-Driven Taxonomy: Rationale and Overview} \label{sec:taxonomy-overview}
\begin{figure*}[!ht]
\centering
\resizebox{\textwidth}{!}{
\sf
\begin{tikzpicture}
\node (TAX) {
\begin{forest}
      for tree={
        grow=east,
        reversed=true,
        anchor=base west,
        parent anchor=east,
        child anchor=west,
        base=left,
        font=\normalsize,
        rectangle,
        draw=hidden-draw,
        rounded corners,
        align=left,
        minimum width=4em,
        edge+={black, line width=0.5pt},
        s sep=3pt,
        l sep = 8mm,
        inner xsep=2pt,
        inner ysep=3pt,
        edge path={
          \noexpand\path [draw, \forestoption{edge}]
          (!u.parent anchor) -- ++(3.5mm,0) |- (.child anchor) \forestoption{edge label};
        },
        ver/.style={rotate=90, child anchor=north, parent anchor=south, anchor=center},
      },
      where level=1{text width=12em,font=\small,}{},
      where level=2{text width=12em,font=\small,}{},
      where level=3{text width=13em,font=\small,}{},
      where level=4{text width=21em,font=\small,}{},
      where level=5{text width=22em,font=\small,}{},
      [
            {\includegraphics[width=\myMheight]{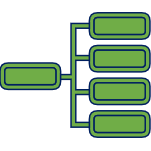}}
            AQA Methods (\cref{sec:taxonomy}), color=black, fill=gray!18, text=black
[
              {\includegraphics[width=\myMheight]{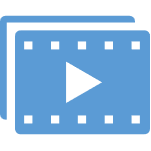}}
              Video-Based (\cref{sec:taxonomy-video_based}), color=black, fill=OfficeBlue!25, text=black
              [
                {\includegraphics[width=\myMheight]{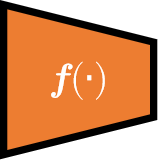}}
                Feature Extraction, color=black, fill=OfficeBlue!12, text=black
                [ 
                {Using fixed backbones mainly for long-term AQA: GDLT \cite{xu2022likert}, PHI \cite{zhou2025phi}
                }, color=black, fill=OfficeBlue!5, text=black, text width=29em
                ]
                [
                {Using tuning backbones:
                TSA-Net \cite{wang2021tsa},
                PECoP \cite{dadashzadeh2024pecop},
                FineParser \cite{xu2024fineparser}
                }, color=black, fill=OfficeBlue!5, text=black, text width=29em
                ]
              ]
              [
                {\includegraphics[width=\myMheight]{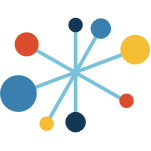}}
                Representation Learning, color=black, fill=OfficeBlue!12, text=black
                [
                {Spatial-aware modeling:
                JRG \cite{pan2019action,pan2021adaptive},
                SAP-Net \cite{gedamu2024self}, 
                FSPN \cite{gedamu2023fine}
                }, color=black, fill=OfficeBlue!5, text=black, text width=29em
                ]
                [
                {Temporal-aware modeling:
                TSA \cite{xu2022finediving},
                HGCN \cite{zhou2023hierarchical},
                FineCausal \cite{han2025finecausal}
                },
                color=black, fill=OfficeBlue!5, text=black, text width=29em
                ]
              ]
              [
                {\includegraphics[width=\myMheight]{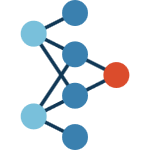}}
                Quality Prediction, color=black, fill=OfficeBlue!12, text=black
                [
                {Direct assessment: 
                USDL \cite{tang2020uncertainty}, 
                DAE-AQA \cite{zhang2024auto},
                CoFInAl \cite{zhou2024cofinal}
                }, color=black, fill=OfficeBlue!5, text=black, text width=29em
                ]
                [
                {Contrastive assessment: 
                CoRe \cite{yu2021group},
                PCLN \cite{li2022pairwise},
                Rhythmer \cite{luo2024rhythmer}
                }, color=black, fill=OfficeBlue!5, text=black, text width=29em
                ]
              ]
            ]
[  
              {\includegraphics[width=\myMheight]{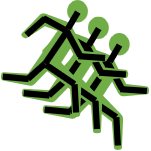}}
              Skeleton-Based (\cref{sec:taxonomy-skeleton_based}), color=black, fill=OfficeOrange!22, text=black
              [
                {\includegraphics[width=\myMheight]{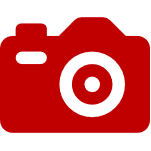}}
                Skeletal Acquisition, color=black, fill=OfficeOrange!12, text=black
                [
                {
                RGB cameras: 
                OpenPose \cite{cao2017realtime},
                MediaPipe \cite{lugaresi2019mediapipe},
                ViTPose \cite{xu2022vitpose}
                }, color=black, fill=OfficeOrange!5, text=black, text width=29em
                ]
                [
                {
                Depth cameras: 
                Microsoft Kinect v1/v2 (e.g., EGCN \cite{bruce2022egcn}, EGCN++ \cite{bruce2024egcn++})
                }, color=black, fill=OfficeOrange!5, text=black, text width=29em
                ]
                [
                {
                Marker-based MoCap systems: 
                Qualisys \cite{wang2022skeleton}, Vicon \cite{bruce2021skeleton}
                }, color=black, fill=OfficeOrange!5, text=black, text width=29em
                ]
              ]
              [
                {\includegraphics[width=\myMheight]{representation.png}}
                Representation Learning, color=black, fill=OfficeOrange!12, text=black
                [
                {
                Basic methods: 
                DCT+SVC \cite{wang2020towards},
                CNN+LSTM \cite{liao2020deep},
                ST-GCN \cite{yan2018spatial}
                }, color=black, fill=OfficeOrange!5, text=black, text width=29em
                ]
                [
                {
                Robust mechanisms: 
                EGCN \cite{bruce2022egcn},
                EGCN++ \cite{bruce2024egcn++}, 
                AAST-GCN \cite{zhou2024attention}
                }, color=black, fill=OfficeOrange!5, text=black, text width=29em
                ]
              ]
            ]
[ 
              {\includegraphics[width=\myMheight]{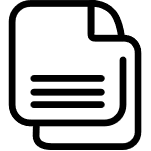}}
              $\cdots$
              Multi-Modal (\cref{sec:taxonomy-multi_modality}), color=black, fill=OfficeGreen!22, text=black
              [
                {\includegraphics[width=\myMheight]{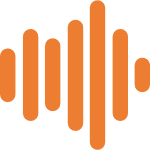}}
                Audio-Assisted, color=black, fill=OfficeGreen!12, text=black
                [
                {
                PISA \cite{parmar2021piano},
                MLP-Mixer \cite{xia2023skating},
                Dance-AQA \cite{zhong2023contrastive},
                PAMFN \cite{zeng2024multimodal},
                LMAC-Net \cite{wang2025attentiondriven}
                }, color=black, fill=OfficeGreen!5, text=black, text width=29em
                ]
              ]
              [ 
                {\includegraphics[width=\myMheight]{text.png}}
                Text-Assisted, color=black, fill=OfficeGreen!12, text=black
                [
                {
                SGN-AQA \cite{du2023learning},
                NAE-AQA \cite{zhang2024narrative},
                MGVLA \cite{xu2024vision},
                MLAVL \cite{xu2025language},
                }, color=black, fill=OfficeGreen!5, text=black, text width=29em
                ]
              ]
              [ 
                {\includegraphics[width=\myMheight]{video.png}} {\includegraphics[width=\myMheight]{skeleton.png}}
                Video-Skeleton, color=black, fill=OfficeGreen!12, text=black
                [
                {
                LUSD-Net \cite{ji2023localization},
                AGF-Olympics \cite{zahan2024learning},
                2M-AF \cite{ding20242m}, 
                LucidAction \cite{dong2025lucidaction}
                }, color=black, fill=OfficeGreen!5, text=black, text width=29em
                ]
              ]
              [ 
                {\includegraphics[width=\myMheight]{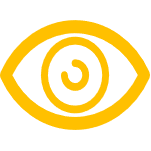}}
                {\includegraphics[width=\myMheight]{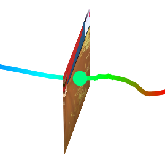}}
                Others-Assisted, color=black, fill=OfficeGreen!12, text=black
                [
                {
                MRehab \cite{khan2022mrehab},
                SASR \cite{kim2024automatic},
                RAAN-TL/RN \cite{huang2024egoexolearn},
                UEF-AQA \cite{kim2024automatic}
                }, color=black, fill=OfficeGreen!5, text=black, text width=29em
                ]
              ]
            ]
      ]
\end{forest}
};
\node[fill=none, font=\small, align=center]
  at ($(TAX.south west)+(2cm,3.8cm)$)
  {
    \faBullseye\ Rationale:\\[2pt]
    \textbf{Input modality determines}\\ \textbf{model selection}
  };

\node[fill=none, font=\small, align=center]
  at ($(TAX.south west)+(7cm,5.6cm)$)
  {
    \color{OfficeOrange}
    \faBullseye\ Rationale:\\[2pt]
    \color{OfficeOrange}\textbf{High-cost data acquisition}
  };

\node[fill=none, font=\small, align=center, text=OfficeBlue]
  at ($(TAX.south west)+(7cm,7.3cm)$)
  {
    \faBullseye\ Rationale:\\[2pt]
    \textbf{Redundant frames and sparse}\\
    \textbf{fine-grained cues complicate}\\
    \textbf{feature extraction/aggregation}
  };

\node[fill=none, font=\small, align=center, text=OfficeGreen]
  at ($(TAX.south west)+(7cm,2.8cm)$)
  {
    \faBullseye\ Rationale:\\[2pt]
    \textbf{Complementary cues}\\
    \textbf{modality selection \& fusion}
  };

\end{tikzpicture}
}
\caption{Our hierarchical taxonomy organized by input modalities and key method design rationales.}
\label{fig:taxonomy}
\vspace{-0.25cm} 
\end{figure*}

As illustrated in \cref{fig:taxonomy}, we organize AQA methods using a modality-driven taxonomy because the input modality largely determines the available motion cues, data representation, and modeling pipeline. 
At the \textbf{first} level, methods are divided into video-based, skeleton-based, and multi-modal branches. Video-based and skeleton-based AQA form two primary paradigms, as they capture motion through visual appearance and structured joint trajectories, respectively. 
At the \textbf{second} level, each branch is further refined according to its core challenges. Video-based AQA focuses on extracting subtle cues from long and redundant visual sequences, skeleton-based AQA emphasizes robust modeling under noisy pose data and costly acquisition, while multi-modal AQA centers on effectively fusing complementary modalities. 
In this survey, we distinguish paradigm-level advances from incremental improvements. 
The former introduces new modeling formulations (e.g., modality shifts or supervision paradigms), while the latter mainly refines existing pipelines through architectural modifications.

\subsection{Video-Based Approach} \label{sec:taxonomy-video_based}
Video-based AQA methods exploit rich RGB visual information and the wide availability of video data to capture fine-grained motion and contextual cues, making them the dominant paradigm in gymnastics~\cite{zeng2020hybrid,lin2025enhancing}, diving~\cite{parmar2019action,parmar2019and}, etc. However, assessing action quality from raw videos faces intrinsic challenges, including long and redundant temporal sequences versus limited computational resources \cite{parmar2017learning,xu2022likert}, sparse and subtle quality cues easily overwhelmed by background variations \cite{nagai2021action,zhang2025scaled}, coarse features transferred from action recognition with domain shift \cite{zhou2024cofinal,zhou2025phi}, and limited labeled data due to high annotation cost \cite{zhou2024magr,zhang2022semi}. These challenges arise at different stages of the processing pipeline and motivate distinct design choices. Accordingly, we organize video-based AQA methods by \emph{how they address these issues across three stages}: feature extraction, video-level representation learning, and quality prediction.

\subsubsection{Feature Extraction} 
\label{sec:feature_extraction}
As illustrated in \cref{fig:backbone_evolution_2012_2025}, early AQA methods~\cite{pirsiavash2014assessing,zhang2014relative} relied on handcrafted features, which were adequate for small-scale datasets but limited in capturing rich spatiotemporal cues. With the emergence of deep learning, more expressive representations became possible. However, because AQA datasets remain relatively small due to costly annotations, training deep models from scratch is often impractical for long and highly redundant video data. Consequently, most works adopt backbone networks~\cite{krizhevsky2012imagenet,tran2015learning,qiu2017learning,carreira2017quo,liu2022video} pre-trained on large-scale image or action recognition datasets~\cite{krizhevsky2012imagenet,tran2015learning,kay2017kinetics,carreira2018short}. Yet these models are not optimized for the fine-grained requirements of AQA, and the high computational cost of long videos further limits extensive fine-tuning. As a result, recent approaches either adapt the backbone to the AQA domain or refine representations at later stages. In the following, we review commonly used backbones and representative adaptation strategies.

\begin{figure*}[!ht]
    \centering
    \sf
    \resizebox{0.9\linewidth}{!}{
    \begin{tikzpicture}[
      font=\scriptsize,
      axis/.style={very thick, draw=gray!65, -{Stealth[length=3.6mm,width=2.2mm]}},
      yearblock/.style={rounded corners=1.5pt},
      yearlabel/.style={text=black, font=\scriptsize\bfseries},
      bbmark/.style={circle, draw=none, inner sep=1.25pt},
      bbicon/.style={draw=none, inner sep=1.25pt, font=\scriptsize},
      bbtext/.style={align=center, inner sep=1pt},
      papermark/.style={circle, draw=none, inner sep=1.15pt},
      paperbox/.style={
        rounded corners=2pt,
        inner xsep=3pt, inner ysep=2pt,
        fill=gray!6, draw=none,
        align=center,
        text width=2.5em,
        font=\scriptsize
      },
    ]
    
\providecolor{OfficeBlue}{rgb}{0.00,0.45,0.74}    \providecolor{OfficeOrange}{rgb}{0.85,0.33,0.10}  \definecolor{OfficeYellow}{rgb}{0.93,0.69,0.13}   \providecolor{OfficeGreen}{rgb}{0.25,0.60,0.35}   \providecolor{OfficePurple}{rgb}{0.49,0.18,0.56}  \definecolor{FigureBG}{RGB}{245,248,252}           \definecolor{OfficePurple}{rgb}{0.49,0.18,0.56}

\def\YearStart{2012}
    \def\YearEnd{2025}
    \pgfmathsetmacro{\Xmax}{\YearEnd-\YearStart}
\def\ActiveYears{2014,2017,2018,2019,2020,2021,2022,2023,2024,2025}
    \def\NodeR{0.30} 

\fill[FigureBG, rounded corners=0pt] (-0.55,-2.2) rectangle (\Xmax+1.6,1.5);
    
\def\TopY{1.00}          \def\PaperBaseY{-0.6}   \def\PaperStepY{0.32}    \def\DotToBoxGap{0.18}   

\newcommand{\NextPaperOffset}[1]{\ifcsname paperstack@#1\endcsname\else
        \expandafter\gdef\csname paperstack@#1\endcsname{0}\fi
      \edef\PaperOffset{\csname paperstack@#1\endcsname}\expandafter\xdef\csname paperstack@#1\endcsname{\number\numexpr\PaperOffset+1\relax}}
\makeatletter
    \newcommand{\maybecite}[1]{\@ifundefined{b@#1}{}{~\cite{#1}}}
    \makeatother
    
\draw[axis] (0,0) -- (\Xmax+0.80,0);
    \node[anchor=west, text=gray!70] at (\Xmax+0.88,0) {Year};
    
\foreach \y in {\YearStart,...,\YearEnd}{
      \pgfmathsetmacro{\x}{\y-\YearStart}
      \def\isActive{0}
      \foreach \ay in \ActiveYears {\ifnum\ay=\y\relax\global\def\isActive{1}\fi
      }
      \ifnum\isActive=1
        \def\NodeFill{OfficeYellow!70}\def\NodeHalo{OfficeYellow!30}\def\LabelColor{black}
      \else
        \def\NodeFill{gray!25}\def\NodeHalo{gray!16}\def\LabelColor{black}
      \fi
      \path[fill=\NodeHalo, draw=none] (\x,0) circle[radius=\NodeR+0.08];
      \path[fill=\NodeFill, draw=none] (\x,0) circle[radius=\NodeR];
      \node[yearlabel, text=\LabelColor] at (\x,0) {\y};
    }
    
\def\Backbones{
      2012/OfficeBlue/{\faStar}/{AlexNet}/alexnet2012,
      2016/OfficeBlue/{\faSitemap}/{ResNet}/he2016deep,
      2015/OfficeOrange/{\faCubes}/{C3D}/tran2015c3d,
      2017/OfficeOrange/{\faFilm}/{I3D}/carreira2017quo,
      2022/OfficeGreen/{\faGears}/{VST}/wang2022vst
    }
    
    \foreach \yy/\cc/\ico/\name/\citekey in \Backbones {
      \pgfmathsetmacro{\xx}{\yy-\YearStart}
      \draw[color=black, dash pattern=on 2pt off 1.2pt] (\xx,0.24) -- (\xx,0.62);
      \node[bbtext, text=\cc] at (\xx,\TopY) {\ico\\{\name}};
    }
\pgfmathsetmacro{\xsrc}{2017-\YearStart}
    \pgfmathsetmacro{\xtgt}{2018-\YearStart}
    \draw[color=black, dash pattern=on 2pt off 1.2pt] (\xsrc,0.24) -- (\xtgt,0.62);
    \node[bbtext, text=OfficeOrange] at (\xtgt,\TopY) {\faRandom\\{P3D}};
    
\def\Papers{
      2014/OfficePurple/{\faCogs~\cite{zhang2014relative}},
      2014/OfficePurple/{\faWrench~\cite{pirsiavash2014assessing}},
      2017/OfficeOrange/{\faCubes~\cite{parmar2017learning}},
      2018/OfficeOrange/{\faRandom~\cite{xiang2018s3d}},
      2018/OfficeBlue/{\faStar~\cite{doughty2018s}},
      2021/OfficeOrange/{\faRandom~\cite{dong2021learning}},
      2023/OfficeOrange/{\faRandom~\cite{zhang2023label}},
      2019/OfficeOrange/{\faFilm~\cite{parmar2019and}},
      2019/OfficeOrange/{\faFilm~\cite{pan2019action}},
      2019/OfficeBlue/{\faSitemap~\cite{li2019manipulation}},
      2020/OfficeOrange/{\faFilm~\cite{tang2020uncertainty}},
      2020/OfficeOrange/{\faFilm~\cite{zeng2020hybrid}},
      2021/OfficeOrange/{\faFilm~\cite{yu2021group}},
      2021/OfficeOrange/{\faFilm~\cite{wang2021tsa}},
      2022/OfficeGreen/{\faGears~\cite{xu2022likert}},
      2022/OfficeOrange/{\faFilm~\cite{bai2022action}},
      2022/OfficeOrange/{\faFilm~\cite{pan2021adaptive}},
      2022/OfficeBlue/{\faSitemap~\cite{li2022pairwise}},
      2023/OfficeOrange/{\faFilm~\cite{gao2023automatic}},
      2023/OfficeOrange/{\faFilm~\cite{zhou2023hierarchical}},
      2023/OfficeOrange/{\faFilm~\cite{gedamu2023fine}},
      2024/OfficeGreen/{\faGears~\cite{zhou2024cofinal}},
      2024/OfficeOrange/{\faFilm~\cite{dadashzadeh2024pecop}},
      2024/OfficeOrange/{\faFilm~\cite{zhang2024auto}},
      2024/OfficeOrange/{\faFilm~\cite{ke2024two}},
      2025/OfficeGreen/{\faGears~\cite{zhou2025phi}},
      2025/OfficeGreen/{\faGears~\cite{han2025caflow}},
      2025/OfficeOrange/{\faFilm~\cite{luo2024rhythmer}},
      2025/OfficeOrange/{\faFilm~\cite{xu2025human}}
    }

\foreach \yy in {\YearStart,...,\YearEnd}{
      \expandafter\gdef\csname paperstack@\yy\endcsname{0}}

    \foreach \yy/\cc/\txt in \Papers {
      \pgfmathsetmacro{\xx}{\yy-\YearStart}
      \NextPaperOffset{\yy}

\pgfmathsetmacro{\py}{\PaperBaseY - (\PaperOffset*\PaperStepY)}
\ifnum\PaperOffset=0
        \draw[color=black, dash pattern=on 2pt off 1.2pt] (\xx,-\NodeR) -- (\xx,\py-0.05);
      \fi
      \node[paperbox, anchor=north,fill=\cc!8] at (\xx,\py-\DotToBoxGap) {\color{\cc}\txt};
    }
    
\node[bbmark, color=OfficeBlue, fill=OfficeBlue,  label={right:2D CNN}]      at (12.8,1.25) {};
    \node[bbmark, color=OfficeOrange, fill=OfficeOrange,label={right:3D CNN}]      at (12.8,1.00) {};
    \node[bbmark, color=OfficeGreen, fill=OfficeGreen, label={right:Transformer}] at (12.8,0.75) {};
    \node[bbicon, text=OfficePurple, label={right:{HoG }}] at (-0.2,-1.55) {\faCogs};
    \node[bbicon, text=OfficePurple, label={right:{Gabor }}] at (-0.2,-1.80) {\faWrench};
    \end{tikzpicture}
    }
    \caption{Timeline of backbone architectures (top) and representative video-based AQA papers (bottom).}
    \label{fig:backbone_evolution_2012_2025}
    \vspace{-0.25cm}
\end{figure*}

\myPara{Representative Backbones in AQA}
As summarized in \cref{tab:backbone,fig:backbone_evolution_2012_2025}, commonly used backbones in video-based AQA can be broadly grouped into three categories: 2D CNNs, 3D CNNs, and transformer-based models, reflecting different inductive biases for modeling action dynamics. 
Early studies typically employed lightweight 2D CNNs, such as AlexNet~\cite{krizhevsky2012imagenet} (e.g.,~\cite{doughty2018s}) and ResNet~\cite{he2016deep} (e.g.,~\cite{li2019manipulation,li2022pairwise}), whose strong spatial inductive bias enables efficient appearance modeling but lacks explicit temporal reasoning. 
To better capture motion dynamics, subsequent works increasingly adopted 3D CNNs, including C3D~\cite{tran2015learning} (e.g.,~\cite{parmar2017learning,parmar2019action}), P3D~\cite{qiu2017learning} (e.g.,~\cite{xiang2018s3d}), and especially I3D~\cite{carreira2017quo} (e.g.,~\cite{pan2019action,bai2022action,xu2024fineparser,zhou2023hierarchical}), which introduce a local spatiotemporal inductive bias through 3D convolutions and have become the dominant choice for short- to medium-length AQA tasks. 
More recently, transformer-based backbones such as VST~\cite{liu2022video}, adopted in~\cite{xu2022likert,du2023learning,zhou2024cofinal,dong2024interpretable}, relax these locality assumptions and rely on global attention to model long-range temporal dependencies, making them particularly suitable for long-term AQA, albeit at a substantially higher computational cost.

\begin{table*}
    \rowcolors{2}{gray!3}{gray!12}
    \centering
    \caption{Representative backbone architectures for video-based AQA and their characteristics.}
    \resizebox{\linewidth}{!}{
    \begin{tabular}{rccrm{5cm}m{4cm}m{4cm}cccc}
    \toprule
    \textbf{Backbone} & \textbf{Type} & \textbf{Typical Works} & \textbf{Pre-Trained Data} & \textbf{Spatial-Temporal Modeling} & \textbf{Efficiency} & \textbf{Generalization} \\
    \midrule
    AlexNet \cite{krizhevsky2012imagenet}   & 2D CNN
    & \cite{doughty2018s}
    & ImageNet \cite{krizhevsky2012imagenet}
    &
    \faStar\faStarO\faStarO\faStarO\faStarO:\par Strong spatial appearance; no explicit temporal modeling.
    &
    \faStar\faStar\faStar\faStar\faStar\par Lightweight and computationally efficient
    &
    \faStar\faStar\faStar\faStarO\faStarO\par Limited cross-domain generalization
    \\
    ResNet \cite{he2016deep} & 2D CNN
    & \cite{li2019manipulation,li2022pairwise} & ImageNet \cite{krizhevsky2012imagenet}
    &
    \faStar\faStar\faStarO\faStarO\faStarO\par Strong spatial features; temporal modeling requires extra modules
    &
    \faStar\faStar\faStar\faStar\faStarO\par Efficient with deeper representations
    &
    \faStar\faStar\faStar\faStar\faStarO\par Good transferability with fine-tuning
    \\
    C3D \cite{tran2015learning} & 3D CNN  & \cite{parmar2017learning,parmar2019action} & Sports1M \cite{tran2015learning}
    &
    \faStar\faStar\faStar\faStar\faStarO\par Joint spatiotemporal modeling via 3D convolutions
    &
    \faStar\faStar\faStarO\faStarO\faStarO\par High computation and memory cost
    &
    \faStar\faStar\faStar\faStarO\faStarO\par Moderate transferability
    \\
    P3D \cite{qiu2017learning} & 3D CNN  & \cite{xiang2018s3d,zhang2023label,dong2021learning} & Kinetics-400 \cite{carreira2017quo}
    &
    \faStar\faStar\faStar\faStarO\faStarO\par Factorized spatial and temporal convolutions
    &
    \faStar\faStar\faStar\faStar\faStarO\par More efficient than full 3D CNNs
    &
    \faStar\faStar\faStar\faStarO\faStarO\par Moderate transferability
    \\
    I3D \cite{carreira2017quo} & 3D CNN  & \cite{pan2019action,bai2022action,xu2024fineparser,zhou2023hierarchical} & Kinetics-400 \cite{carreira2017quo}
    &
    \faStar\faStar\faStar\faStar\faStar\par Strong spatiotemporal modeling via inflated 2D filters
    &
    \faStar\faStar\faStarO\faStarO\faStarO\par Computationally heavy due to 3D operations
    &
    \faStar\faStar\faStar\faStar\faStarO\par Strong transferability with fine-tuning
    \\
    VST \cite{liu2022video} & Transformer & \cite{xu2022likert,du2023learning,zhou2024cofinal,dong2024interpretable} & Kinetics-600 \cite{carreira2018short}
    &
    \faStar\faStar\faStar\faStar\faStar\par Strong long-range temporal modeling with global attention
    &
    \faStar\faStarO\faStarO\faStarO\faStarO\par Expensive due to self-attention
    &
    \faStar\faStar\faStar\faStar\faStarO\par Good generalization in complex scenarios
    \\
    \bottomrule
    \end{tabular}
    }
    \label{tab:backbone}
\end{table*}

\myPara{Efficiency, Transferability, and Practical Choices}
The different inductive biases of 2D CNNs, 3D CNNs, and transformers lead to distinct trade-offs in efficiency and temporal modeling. Compared with 2D CNNs, 3D CNNs incur substantially higher computational and memory costs, while transformer-based models further increase this burden due to global attention. To make training tractable, many studies adopt temporal down-sampling strategies, such as key clip selection~\cite{li2018scoringnet} or uniform clip partitioning, although this may discard subtle yet critical cues and weaken long-range temporal coherence. 
Meanwhile, domain shift remains a fundamental issue. Backbones pre-trained for action recognition often encode coarse semantics that are insufficient for the fine-grained requirements of AQA~\cite{zhou2024cofinal,dadashzadeh2024pecop,zhou2025phi}. 
In practice, a clear trade-off emerges: 2D CNNs favor efficiency, 3D CNNs balance spatiotemporal modeling and cost for moderate-length actions, while transformer-based backbones are more suitable for long and complex sequences that require long-range temporal reasoning.

\myPara{Backbone Tuning Strategies}
To mitigate the domain gap between pre-trained and AQA tasks, full fine-tuning is the most straightforward approach. However, it is prone to overfitting on small-scale AQA datasets~\cite{zhou2025continual}. Consequently, several works explore constrained backbone adaptation strategies that improve task relevance while limiting parameter updates. 
For example, TSA-Net~\cite{wang2021tsa} injects human-centric masks into I3D to emphasize critical regions with minimal overhead (see \cref{fig:fine-grained-modeling-a}), and this idea is further extended by FineParser~\cite{xu2024fineparser} and Uni-FineParser~\cite{xu2025human}. 
In contrast, PECoP~\cite{dadashzadeh2024pecop} introduces lightweight 3D components together with self-supervised objectives to learn in-domain spatiotemporal features, and then freezes the adapted backbone to reduce overfitting risk in downstream AQA tasks. 
These strategies offer a practical compromise between transferability and efficiency. For \emph{long-term AQA}~\cite{xu2022likert,du2023learning,zhou2024cofinal,dong2024interpretable}, where minute-level videos are processed, directly fine-tuning the backbone is often impractical due to prohibitive computational cost. Such scenarios are therefore more commonly addressed by feature adaptation at later stages (see \cref{sec:representaion-learning}).

\begin{figure*}[!ht]
    \centering
    \sf
    \resizebox{0.8\linewidth}{!}{
    \begin{tikzpicture}[
      font=\scriptsize,
      axis/.style={very thick, draw=gray!65, -{Stealth[length=3.2mm,width=2mm]}},
      yearblock/.style={rounded corners=1.5pt},
      yearlabel/.style={font=\scriptsize\bfseries},
      paperbox/.style={
        rounded corners=2pt,
        inner xsep=3pt, inner ysep=2pt,
        draw=none,
        align=center,
        text width=2.6em,
        font=\scriptsize
      },
    ]

\definecolor{SpatialColor}{rgb}{0.00,0.45,0.74}   \definecolor{TemporalColor}{rgb}{0.85,0.33,0.10}  \definecolor{OfficeYellow}{rgb}{0.93,0.69,0.13}
\definecolor{FigureBG}{RGB}{245,248,252}

\def\YearStart{2014}
    \def\YearEnd{2025}
    \pgfmathsetmacro{\Xmax}{\YearEnd-\YearStart}
    \def\ActiveYears{2017,2019,2020,2021,2022,2023,2024,2025}
    \def\NodeR{0.30}      

\path[fill=FigureBG, draw=none, rounded corners=3pt]
      (-0.65,-2.25) rectangle (\Xmax+1.75,2.25);

\def\SpatialBaseY{0.68}
    \def\TemporalBaseY{-0.68}
    \def\PaperStepY{0.34}

\newcommand{\AqaNextS}[1]{\ifcsname aqa@s@#1\endcsname\else\expandafter\gdef\csname aqa@s@#1\endcsname{0}\fi
      \edef\SOff{\csname aqa@s@#1\endcsname}\expandafter\xdef\csname aqa@s@#1\endcsname{\number\numexpr\SOff+1\relax}}
    \newcommand{\AqaNextT}[1]{\ifcsname aqa@t@#1\endcsname\else\expandafter\gdef\csname aqa@t@#1\endcsname{0}\fi
      \edef\TOff{\csname aqa@t@#1\endcsname}\expandafter\xdef\csname aqa@t@#1\endcsname{\number\numexpr\TOff+1\relax}}

\draw[line width=2.2pt, draw=gray!20, line cap=round]
      (0,-0.015) -- (\Xmax+0.78,-0.015);

\draw[axis, line cap=round]
      (0,0) -- (\Xmax+0.80,0);

\foreach \xx in {0.6,1.6,...,\Xmax} {
      \draw[gray!45, line width=0.55pt]
        (\xx-0.08,0.07) -- (\xx,0) -- (\xx-0.08,-0.07);
    }

    \node[anchor=west, text=gray!70, font=\scriptsize\bfseries]
      at (\Xmax+0.92,0) {Year};

\foreach \y in {\YearStart,...,\YearEnd}{
      \pgfmathsetmacro{\x}{\y-\YearStart}
      \def\isActive{0}
      \foreach \ay in \ActiveYears { \ifnum\ay=\y\relax\global\def\isActive{1}\fi }
      \ifnum\isActive=1
        \def\Fill{OfficeYellow!68}\def\Halo{OfficeYellow!28}\def\TextCol{black}
      \else
        \def\Fill{gray!25}\def\Halo{gray!16}\def\TextCol{black}
      \fi
      \path[fill=\Halo, draw=none] (\x,0) circle[radius=\NodeR+0.08];
      \path[fill=\Fill, draw=none] (\x,0) circle[radius=\NodeR];
      \node[yearlabel, text=\TextCol] at (\x,0) {\y};
    }

\def\SpatialPapers{
    2019/{\faSitemap}/pan2019action,
    2022/{\faSitemap}/pan2021adaptive,
    2019/{\faSitemap}/li2019manipulation,
    2021/{\faCube}/nagai2021action,
    2023/{\faCube}/gedamu2023fine,
    2025/{\faCube}/zhang2025scaled,
    2024/{\faCube}/gedamu2024self,
    2022/{\faSitemap}/li2022surgical,
    2024/{\faSitemap}/okamoto2024hierarchical,
    2020/{\faSitemap}/gao2020asymmetric,
    2023/{\faSitemap}/gao2023automatic,
    2023/{\faSitemap}/zhang2023logo}

\def\TemporalPapers{
    2017/{\faListAlt}/parmar2017learning,
    2019/{\faListAlt}/parmar2019action,
    2020/{\faListAlt}/tang2020uncertainty,
    2021/{\faListAlt}/yu2021group,
    2022/{\faListAlt}/xu2022likert,
    2022/{\faListAlt}/bai2022action,
    2023/{\faListAlt}/zhou2023hierarchical,
    2024/{\faListAlt}/dong2024interpretable,
    2023/{\faListAlt}/zhang2023label,
    2022/{\faListOl}/xu2022finediving,
    2024/{\faListOl}/xu2024fineparser,
    2025/{\faListOl}/xu2025human,
    2024/{\faListOl}/xu2024procedure,
    2024/{\faListOl}/he2024collaborative,
    2025/{\faListOl}/he2024achieving}

\foreach \yy in {\YearStart,...,\YearEnd}{
      \expandafter\gdef\csname aqa@s@\yy\endcsname{0}
      \expandafter\gdef\csname aqa@t@\yy\endcsname{0}
    }

\foreach \yy/\ico/\citekey in \SpatialPapers {
      \pgfmathsetmacro{\xx}{\yy-\YearStart}
      \AqaNextS{\yy}
      \pgfmathsetmacro{\py}{\SpatialBaseY + (\SOff*\PaperStepY)} \ifnum\SOff=0
        \draw[color=gray!50, line width=0.45pt, dash pattern=on 2.2pt off 1.6pt]
          (\xx,\NodeR) -- (\xx,\py-0.08);
      \fi
      \node[paperbox, fill=SpatialColor!10, text=SpatialColor, anchor=south]
        at (\xx,\py) {\ico~\cite{\citekey}};
    }

\foreach \yy/\ico/\citekey in \TemporalPapers {
      \pgfmathsetmacro{\xx}{\yy-\YearStart}
      \AqaNextT{\yy}
      \pgfmathsetmacro{\py}{\TemporalBaseY - (\TOff*\PaperStepY)} \ifnum\TOff=0
        \draw[color=gray!50, line width=0.45pt, dash pattern=on 2.2pt off 1.6pt]
          (\xx,-\NodeR) -- (\xx,\py+0.08);
      \fi
      \node[paperbox, fill=TemporalColor!10, text=TemporalColor, anchor=north]
        at (\xx,\py) {\ico~\cite{\citekey}};
    }

\node[anchor=west, text=SpatialColor]  at (-0.55,1.95) {\faCube\ \textbf{Spatial}: Distillation-based refinement};
    \node[anchor=west, text=SpatialColor]  at (-0.55,1.65) {\faSitemap\ \textbf{Spatial}: Cue-driven feature refinement};

    \node[anchor=west, text=TemporalColor] at (-0.55,-1.65) {\faListAlt\ \textbf{Temporal}: Implicit modeling};
    \node[anchor=west, text=TemporalColor] at (-0.55,-1.95) {\faListOl\ \textbf{Temporal}: Explicit modeling};

    \end{tikzpicture}
    }
    \caption{Timeline of representative video-based AQA methods in representation learning, grouped into spatial-aware (above the axis) and temporal-aware (below the axis) strategies over time.}
    \label{fig:representation_timeline}
    \vspace{-0.25cm}
\end{figure*}

\subsubsection{Video-Based Representation Learning} \label{sec:representaion-learning}
Although powerful backbones provide strong clip-level features, feature extraction (see \cref{sec:feature_extraction}) in AQA still faces two fundamental limitations: 
(1) to control computational cost, long videos are usually processed as short clips, which breaks temporal continuity and weakens long-range dependency modeling; and 
(2) features transferred from action recognition are often coarse and domain-shifted, and fine-tuning them is impractical for \emph{long-term AQA} with minute-level videos due to prohibitive cost. 
Video-level representation learning is therefore introduced as an intermediate stage between feature extraction and quality prediction, aiming to refine clip-level features into more quality-aware representations and aggregate them into a coherent video-level embedding that preserves global temporal context. By explicitly modeling finer spatial and temporal cues and integrating information across clips, this stage mitigates both coarse semantics and fragmented temporal modeling. 
Most existing methods address both aspects to some extent, but differ in their primary emphasis. As summarized in \cref{fig:representation_timeline}, representative works are reviewed below and categorized into \emph{spatial-aware} approaches~\cite{gedamu2023fine,gedamu2024self,nagai2021action,li2022surgical}, which enhance human- and object-centric cues, and \emph{temporal-aware} approaches~\cite{parmar2019action,bai2022action,fang2023end,lei2021temporal}, which focus on procedural structure and long-range dependencies.

\begin{figure}[!t]
    \centering
    \includegraphics[width=\linewidth,trim=0 140 0 140,clip]{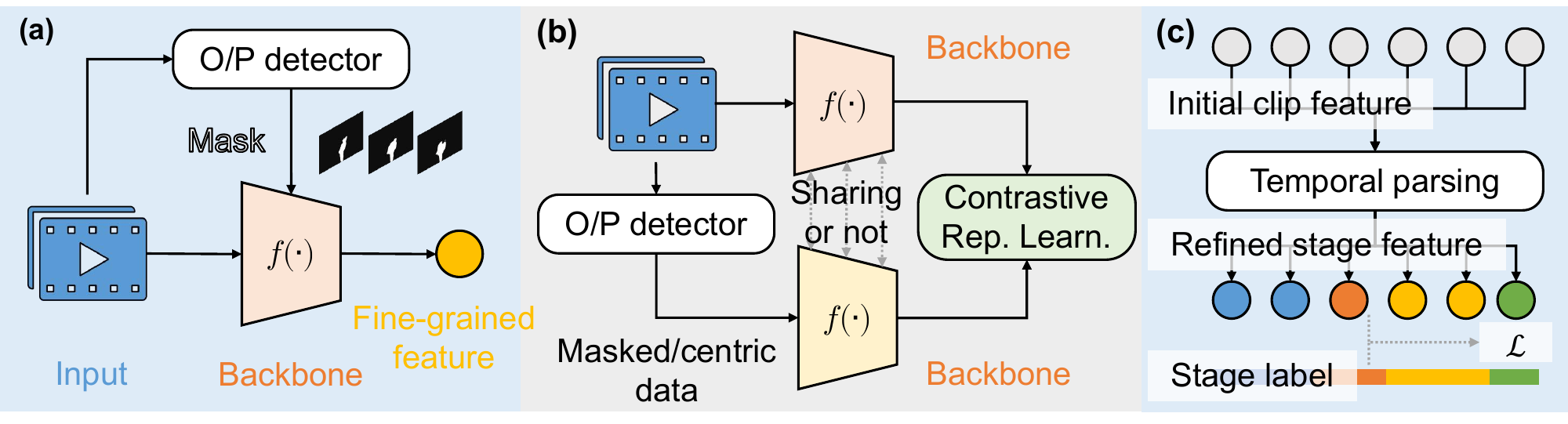}
    \caption{
        Three representative fine-grained reasoning paradigms in AQA.
        (a) Spatial reasoning injects object- or pose-centric masks into backbones to emphasize critical regions.
        (b) Siamese reasoning contrasts raw and masked inputs to highlight object-centric cues, sometimes using unshared branches for richer representations.
        (c) Temporal reasoning models procedural structure by parsing subactions, either implicitly or explicitly.
    }
    \label{fig:fine-grained-modeling}
    \phantomsubcaption\label{fig:fine-grained-modeling-a}\phantomsubcaption\label{fig:fine-grained-modeling-b}\phantomsubcaption\label{fig:fine-grained-modeling-c}\vspace{-0.25cm}
\end{figure}

\myPara{Spatial-Aware Modeling}
\textit{Spatial-aware} modeling aims to emphasize human- and object-centric regions while suppressing background distractions, since subtle quality differences in AQA are often reflected in localized body parts, tools, or interactions rather than global appearance. 
Instead of explicitly modifying the backbone with masks (e.g., \cref{fig:fine-grained-modeling-a}), most works implicitly refine spatial representations during representation learning to improve flexibility (see \cref{fig:fine-grained-modeling-b}). 
These works mainly fall into two representative directions.

One direction is \textbf{distillation-based refinement}, which guides the model to focus on actor-centric regions through strategies such as human-masked supervision and scene-adversarial objectives~\cite{nagai2021action}, teacher--student consistency with pseudo actor-centric labels~\cite{gedamu2024self}, background swapping augmentation~\cite{zhang2025scaled}, and causality-aware foreground--background modeling~\cite{han2025finecausal}. 
These strategies suppress background bias and transfer fine-grained spatial knowledge to the student network, enabling more discriminative representations for accurate AQA.

Another line of work follows \textbf{cue-driven feature refinement}, where spatial representations are enriched by integrating auxiliary cues, such as frame-wise spatial attention to highlight critical regions~\cite{li2019manipulation}, semantic component modeling in surgical scenes~\cite{li2022surgical}, body-part graph reasoning based on pose estimation~\cite{pan2019action,pan2021adaptive}, and interaction relationship ~\cite{gao2020asymmetric,gao2023automatic} or group structure modeling~\cite{zhang2023logo} for multi-agent actions. 
These designs introduce richer semantics into spatial modeling, leading to more interpretable representations of critical regions, while often relying on the quality of auxiliary cues.
Cue-driven designs enrich spatial semantics but depend on external cue quality and add complexity, which may hurt robustness. 
In contrast, distillation-based refinement injects spatial priors implicitly, offering a lightweight and flexible alternative. 
This reveals a key trade-off between explicit semantic structure and implicit, data-driven refinement, motivating designs that balance expressiveness and efficiency.

\begin{figure}[!ht]
    \centering
        \centering
        \includegraphics[width=\linewidth,trim=0 115 0 115,clip]{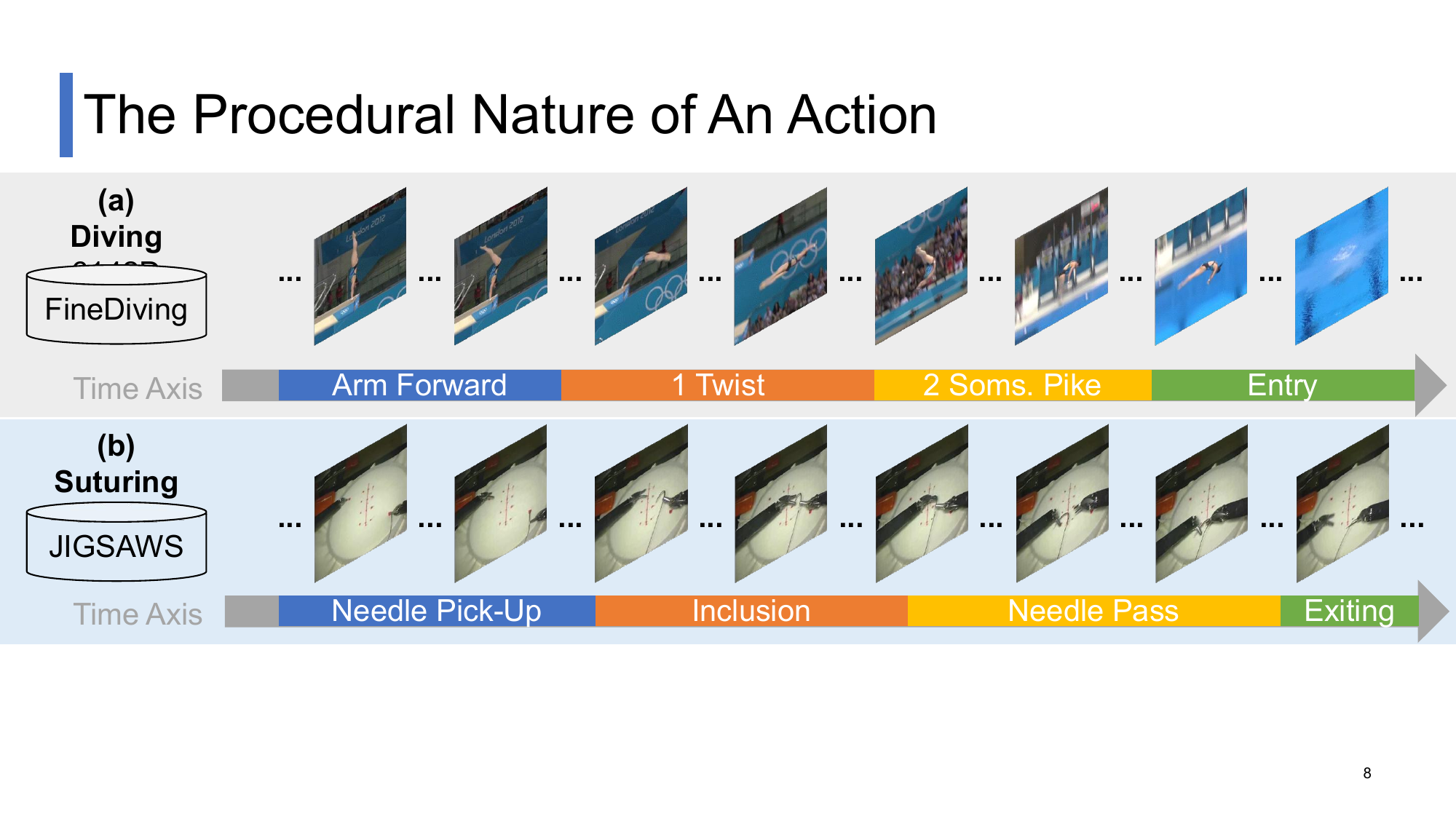}
        \caption{The procedural nature of actions in fine temporal modeling for AQA. (a) illustrates a diving example from the FineDiving dataset~\cite{xu2022finediving}, while (b) depicts a suturing example from the JIGSAWS dataset~\cite{gao2014jhu}.}
        \label{fig:procedure-aqa}
        \phantomsubcaption\label{fig:procedure-aqa-a}\phantomsubcaption\label{fig:procedure-aqa-b}\vspace{-0.25cm}
\end{figure}

\textbf{Temporal-Aware Modeling.}
\textit{Temporal-aware} modeling aims to recover the procedural structure of actions (see \cref{fig:procedure-aqa}) from fragmented clip features and to capture long-range dependencies that are critical for AQA. 
In many AQA settings, an action naturally consists of a sequence of subactions (e.g., approach, takeoff, flight, and entry in \cref{fig:procedure-aqa-a}), and errors in any stage may substantially affect the final score~\cite{xu2022finediving,zhou2023hierarchical,xu2024fineparser,dong2021learning}. 
However, due to efficiency constraints, videos are usually processed as disjoint clips, and naive aggregation (e.g., average pooling~\cite{parmar2017learning}) often leads to vague temporal modeling and loss of contextual cues. 
To address this, existing works mainly fall into two representative directions.

One line of work follows \textbf{implicit temporal modeling}, which learns temporal dependencies directly from data using sequence models. 
Early studies employed LSTMs or TCNs to aggregate clip features~\cite{parmar2019action,wang2020assessing,wang2020towards}, while more recent methods adopt Transformers to better capture long-range interactions~\cite{xu2022likert,bai2022action,fang2023end,lei2021temporal}. 
Typical designs include grade-aware decoding via cross-attention~\cite{xu2022likert}, temporal parsing with learnable queries~\cite{bai2022action}, hierarchical graph reasoning over clips~\cite{zhou2023hierarchical,liu2025adaptive}, multi-stage temporal parsing for subaction decomposition~\cite{dong2021learning,an2024multi,qi2025action}, causality-aware modeling \cite{han2025caflow,han2025finecausal}, and event pattern analysis \cite{liu2021towards,ding2023sedskill}. 
These approaches are flexible and do not require extra annotations, enabling broad applicability across tasks, but often suffer from limited interpretability, higher computational cost, and the risk of shortcut learning or overfitting on small datasets~\cite{dong2024interpretable,dong2026uila}.

In contrast, \textbf{explicit temporal modeling} injects human knowledge of action procedures through temporal annotations. 
Representative works such as TSA and FineParser~\cite{xu2022finediving,xu2024procedure,xu2024fineparser,xu2025human} leverage stage-level supervision to guide feature aggregation and score prediction, leading to more interpretable and fine-grained assessment of each subaction. 
Weakly supervised variants further reduce annotation cost by aligning procedures across videos sharing similar structures~\cite{he2024achieving,he2024collaborative}. 
While these methods offer stronger interpretability and stage-level feedback, they rely heavily on costly temporal annotations, are less scalable, and often generalize poorly to unseen actions. 
Overall, temporal-aware modeling reveals a clear trade-off: implicit approaches favor flexibility and scalability but often lack transparency, whereas explicit designs provide structured and interpretable reasoning at the cost of increased annotation demands and limited generalization. 
These observations suggest that hybrid strategies, which combine data-driven temporal learning with lightweight procedural priors, may offer a promising direction for improving the robustness and practicality of AQA. 

\begin{figure*}[!ht]
    \centering
    \sf
    \resizebox{0.90\linewidth}{!}{
    \begin{tikzpicture}[
      font=\scriptsize,
      axis/.style={very thick, draw=gray!65, -{Stealth[length=3mm,width=2mm]}},
      paper/.style={
        rounded corners=2pt,
        inner xsep=2pt,
        inner ysep=2pt,
        align=center,
        font=\scriptsize,
        text width=2.6em
      },
    ]

\definecolor{DirectColor}{rgb}{0.20,0.60,0.20}      \definecolor{ContrastColor}{rgb}{0.85,0.33,0.10}    \definecolor{YearFill}{rgb}{0.93,0.69,0.13}
    \definecolor{FigureBG}{RGB}{245,248,252}

\def\YearStart{2014}
    \def\YearEnd{2025}
    \pgfmathsetmacro{\Xmax}{\YearEnd-\YearStart}
    \def\ActiveYears{2014,2017,2018,2019,2020,2021,2022,2023,2024,2025}
    \def\NodeR{0.30}

\makeatletter
    \newcommand{\maybecite}[1]{\@ifundefined{b@#1}{}{~\cite{#1}}}
    \makeatother

\path[fill=FigureBG, draw=none, rounded corners=3pt]
      (-0.65,-2.25) rectangle (\Xmax+1.75,2.25);

\draw[line width=2.2pt, draw=gray!20, line cap=round]
      (0,-0.015) -- (\Xmax+0.78,-0.015);
    \draw[axis, line cap=round] (0,0) -- (\Xmax+0.80,0);
    \foreach \xx in {0.6,1.6,...,\Xmax} {
      \draw[gray!45, line width=0.55pt]
        (\xx-0.08,0.07) -- (\xx,0) -- (\xx-0.08,-0.07);
    }
    \node[anchor=west, font=\scriptsize\bfseries, text=gray!70] at (\Xmax+0.92,0) {Year};

\foreach \y in {\YearStart,...,\YearEnd}{
      \pgfmathsetmacro{\x}{\y-\YearStart}
      \def\isActive{0}
      \foreach \ay in \ActiveYears { \ifnum\ay=\y\relax\global\def\isActive{1}\fi }
      \ifnum\isActive=1
        \def\Fill{YearFill!72}\def\Halo{YearFill!28}
      \else
        \def\Fill{gray!25}\def\Halo{gray!16}
      \fi
      \path[fill=\Halo, draw=none] (\x,0) circle[radius=\NodeR+0.08];
      \path[fill=\Fill, draw=none] (\x,0) circle[radius=\NodeR];
      \node[font=\scriptsize\bfseries, text=black] at (\x,0) {\y};
    }

\def\DirectY{0.95}
    \def\ContrastY{-0.95}
    \def\StackStep{0.35}

\newcommand{\NextDirect}[1]{\ifcsname direct@#1\endcsname\else\expandafter\gdef\csname direct@#1\endcsname{0}\fi
      \edef\DOff{\csname direct@#1\endcsname}\expandafter\xdef\csname direct@#1\endcsname{\number\numexpr\DOff+1\relax}}
    \newcommand{\NextContrast}[1]{\ifcsname contrast@#1\endcsname\else\expandafter\gdef\csname contrast@#1\endcsname{0}\fi
      \edef\COff{\csname contrast@#1\endcsname}\expandafter\xdef\csname contrast@#1\endcsname{\number\numexpr\COff+1\relax}}

\foreach \yy/\ico/\ck in {
        2017/{\faPaw}/parmar2017learning,
        2019/{\faPaw}/parmar2019and,
        2025/{\faPaw}/fang2025a,
        2025/{\faEye}/wang2025adaptive,
        2020/{\faEye}/tang2020uncertainty,
        2022/{\faEye}/zhou2022uncertainty,
        2023/{\faEye}/zhou2023hierarchical,
        2024/{\faEye}/zhang2024auto,
        2024/{\faCogs}/matsuyama2023iris,
        2024/{\faCogs}/zhang2023label}{
        \pgfmathsetmacro{\xx}{\yy-\YearStart}
        \NextDirect{\yy}
        \pgfmathsetmacro{\dy}{\DirectY + \DOff*\StackStep} \draw[gray!55, dash pattern=on 2.2pt off 1.6pt, line width=0.45pt] (\xx,\NodeR) -- (\xx,\dy-0.08);
        \node[paper, fill=DirectColor!12, text=DirectColor, anchor=south]
            at (\xx,\dy) {\ico~\cite{\ck}};
    }

\foreach \yy/\ico/\ck in {
        2014/{\faLink}/zhang2014relative,
        2018/{\faLink}/doughty2018s,
        2019/{\faLink}/doughty2019pros,
        2021/{\faLink}/yu2021group,
        2022/{\faReorder}/li2022pairwise,
        2022/{\faReorder}/xu2022finediving,
        2022/{\faReorder}/bai2022action,
        2023/{\faLink}/gedamu2023fine,
        2024/{\faLink}/liu2023figure,
        2024/{\faReorder}/an2024multi,
        2024/{\faReorder}/ke2024two,
        2025/{\faReorder}/qiu2025learning,
        2025/{\faReorder}/luo2024rhythmer,
        2025/{\faLink}/han2025finecausal}{
        \pgfmathsetmacro{\xx}{\yy-\YearStart}
        \NextContrast{\yy}
        \pgfmathsetmacro{\cy}{\ContrastY - \COff*\StackStep} \draw[gray!55, dash pattern=on 2.2pt off 1.6pt, line width=0.45pt] (\xx,-\NodeR) -- (\xx,\cy+0.08);
        \node[paper, fill=ContrastColor!12, text=ContrastColor, anchor=north]
            at (\xx,\cy) {\ico~\cite{\ck}};
    }

\def\LegX{-0.55}
    \node[anchor=west, text=DirectColor, font=\scriptsize]   at (\LegX, 1.80) {\faPaw\ Baseline regression};
    \node[anchor=west, text=DirectColor, font=\scriptsize]   at (\LegX, 1.55) {\faEye\ Uncertainty-aware};
    \node[anchor=west, text=DirectColor, font=\scriptsize]   at (\LegX, 1.30) {\faCogs\ Subaction assessment};
    \node[anchor=west, text=ContrastColor, font=\scriptsize] at (\LegX, -1.75) {\faLink\ Contrastive regression};
    \node[anchor=west, text=ContrastColor, font=\scriptsize] at (\LegX, -2.00) {\faReorder\ Contrastive ranking};
    \end{tikzpicture}
    }
    \caption{Timeline of representative quality prediction methods in AQA, grouped into direct assessment (top) and contrastive assessment (bottom) paradigms.}
    \label{fig:quality_timeline}
\end{figure*}

\vspace{-0.25cm}
 \subsubsection{Quality Prediction}
Quality prediction is the final stage of AQA, which maps video-level representations to assessment outputs and ultimately determines how action quality is quantified. 
From a modeling perspective, this stage differs mainly in whether quality is inferred from an individual action in isolation or from its relation to other actions. 
As visualized in \cref{fig:quality_timeline}, existing methods can therefore be broadly categorized into \emph{direct} and \emph{contrastive} assessments.

\textbf{Direct Assessment.}
\emph{Direct} assessment methods~\cite{parmar2019and,zhang2024auto,zhou2024cofinal,zhou2023hierarchical,li2024continual,zhou2024magr,fang2025a} formulate AQA as regression or classification, predicting absolute scores or grades by minimizing errors with respect to ground truth. 
They are simple to implement and computationally efficient, but often show limited sensitivity and generalization when quality differences are subtle or action contexts vary. 
To alleviate these issues, two representative strategies have been explored.
\textbf{Uncertainty modeling}~\cite{tang2020uncertainty,zhang2024auto,zhou2023hierarchical,ji2023localization} explicitly accounts for ambiguity in human judgments by predicting score distributions rather than deterministic values. 
USDL and MUSDL~\cite{tang2020uncertainty} soften labels with Gaussian distributions for single- and multi-judge settings, while DAE~\cite{zhang2024auto} employs VAE-based modeling to regress scores from latent distributions. 
These approaches improve robustness to noisy annotations, but often rely on assumptions about label variance and may generalize poorly with limited data.
\textbf{Subaction assessment}~\cite{okamoto2024hierarchical,zhang2023label,matsuyama2023iris,lian2023improving} decomposes an action into meaningful stages and predicts sub-scores that are later aggregated into a final score. 
This enables fine-grained diagnosis and improves generalization by leveraging the internal structure of actions. 

\begin{figure}[!ht]
    \centering
    \includegraphics[width=\linewidth,clip,trim=0 95 0 140]{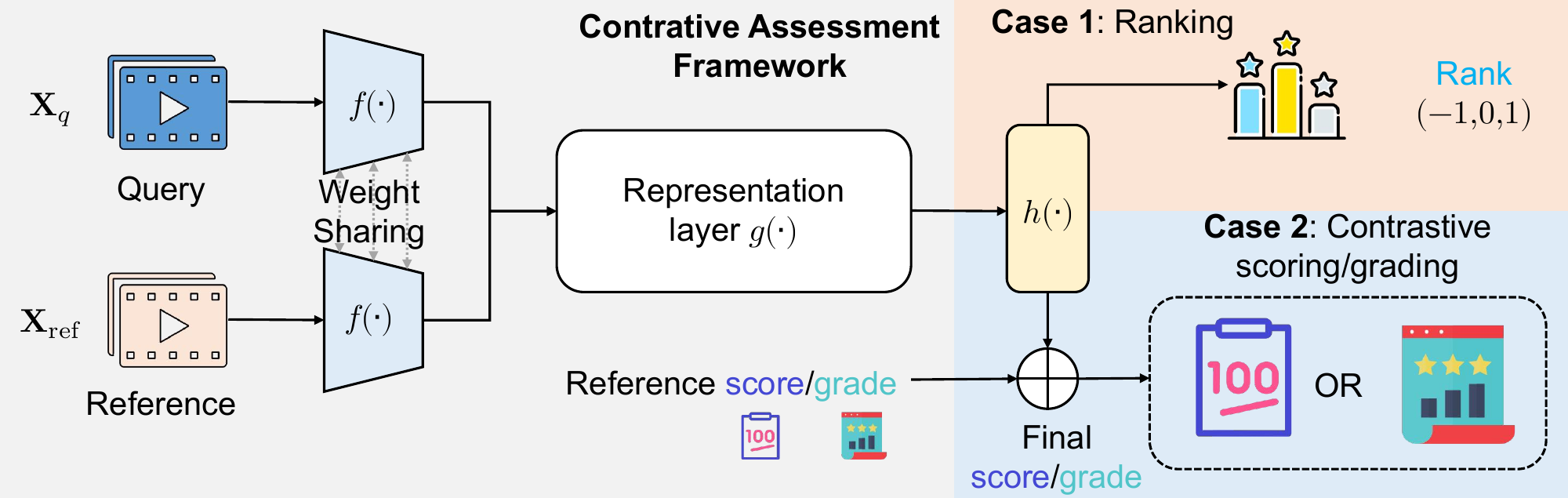}
    \caption{
Pairwise contrastive framework for AQA.
The model compares a target action with a reference action to learn relative quality cues.
Case~1 uses relative ranking (see \cref{eq:ranking}) without exact labels, whereas Case~2 performs contrastive scoring or grading with labeled references.
}
    \label{fig:contrastive-regression-or-ranking}

    \vspace{-0.35cm}
\end{figure}

\myPara{Contrastive Assessment}
\emph{Contrastive assessment} methods~\cite{doughty2018s,yu2021group,xu2022finediving,bai2022action,gedamu2023fine,jain2020action} learn action quality by modeling \emph{relative differences} between actions rather than relying solely on absolute scores. 
Typically implemented with Siamese architectures (see \cref{fig:contrastive-regression-or-ranking}), these methods compare a target video against reference examples, making them particularly sensitive to fine-grained performance variations. 
Originally introduced in ranking-based formulations~\cite{zhang2014relative,doughty2018s,doughty2019pros}, contrastive learning has since been extended to both ranking and scoring settings.
\textbf{Ranking-oriented methods}~\cite{zhang2014relative,doughty2018s,bertasius2017baller,luo2024rhythmer,fang2024better} avoid absolute annotations and instead predict relative orderings between actions. 
This formulation substantially reduces annotation cost and is especially attractive when expert scores are expensive, subjective, or unreliable. 
Such approaches have been successfully applied across domains, including surgical skill assessment \cite{pan2025basket,ding2023sedskill} and teaching quality evaluation \cite{fang2024better,fang2025a}.
\textbf{Contrastive regression methods} incorporate pairwise contrastive learning into score prediction~\cite{yu2021group,xu2022finediving,bai2022action,xu2024procedure,xu2024fineparser,gedamu2023fine,gedamu2024self,jain2020action}. 
Representative designs include group-aware contrastive regression~\cite{yu2021group}, hierarchical contrastive parsing of subactions~\cite{bai2022action}, and multi-view or replay-based contrastive frameworks~\cite{liu2023figure,ke2024two}. 
By explicitly enforcing relative consistency alongside regression objectives, these methods significantly improve sensitivity to subtle quality differences. 

\myPara{Direct and Contrastive Trade-offs}
As summarized in \cref{tab:assessment}, direct and contrastive assessments differ in several ways. 
Direct methods efficiently regress absolute scores but rely on precise annotations and are less sensitive to subtle differences. 
In contrast, contrastive methods learn from relative comparisons, improving fine-grained discrimination and reducing annotation burden, albeit at higher computational cost. 
In practice, direct assessment suits scenarios requiring absolute scoring and real-time inference, while contrastive formulations favor fine-grained differentiation under limited labels. 
This trade-off has motivated hybrid strategies that use relative comparisons during training and absolute score prediction at inference, combining efficient global calibration with fine-grained supervision~\cite{ding2025sceneaware,fang2025a}.

\begin{table*}
    \rowcolors{2}{gray!3}{gray!12}
    \centering
    \caption{Comparison of direct and contrastive assessment methods.}
    \resizebox{\linewidth}{!}{
    \begin{tabular}{>{\hfill}m{2.2cm}m{3.4cm}<{\centering}m{3.6cm}m{4cm}m{4cm}m{4cm}m{4cm}}
    \toprule
    \textbf{Assessment} & \textbf{Typical Works} & \textbf{Easy to Implement} & \textbf{Sensitivity to Variations} & \textbf{Required Data} & 
    \textbf{Generalization} &
    \textbf{Computation} \\
    \midrule
    Direct   &
    \cite{parmar2019action,tang2020uncertainty,zhang2024auto,okamoto2024hierarchical} & 
    \faStar \faStar \faStar \faStar \faStar:\par  Simple
    &
    \faStar \faStar \faStar \faStarO \faStarO:\par 
    Hard to capture details 
    & 
    \faStar \faStar \faStarO \faStarO \faStarO:\par  Less for train
    &
    \faStar \faStar \faStar \faStarO \faStarO:\par
    Moderate 
    &
    \faStar \faStar \faStar \faStar \faStar:\par 
    Efficient
    \\
Contrastive &
    \cite{doughty2018s,doughty2019pros,yu2021group,xu2022finediving,bai2022action,gedamu2024self} &
    \faStar \faStar \faStarO \faStarO \faStarO:\par  Complex pairwise design 
    &
    \faStar \faStar \faStar \faStar \faStar:\par  Robust for subtle differences
    & 
    \faStar\faStar\faStar\faStar\faStar: \par
    More for pairs 
    &
    \faStar \faStar \faStar \faStar \faStarO:\par  Better for relation focus
    &
    \faStar \faStar \faStar \faStarO \faStarO:\par  
    High for reference compute
     \\
    \bottomrule
    \end{tabular}
    }
    \label{tab:assessment}
\end{table*}

\subsection{Skeleton-Based Approach} \label{sec:taxonomy-skeleton_based}

Skeleton-based AQA methods~\cite{kanade2023attention,lei2023multi,wang2022skeleton,li2022skeleton} assess action quality by modeling human joint trajectories in 2D or 3D space, explicitly focusing on body kinematics rather than visual appearance.
Compared with video-based AQA, skeleton-based approaches inherently suppress background variations, making them well-suited for human-centric scenarios with limited object interaction, such as rehabilitation.
However, this abstraction introduces several practical challenges~\cite{pirsiavash2014assessing,wang2021tsa}. First, reliable pose acquisition is difficult in many real-world settings due to occlusions, viewpoint changes, and limited sensing conditions. Second, skeletal representations are inherently sparse and may miss subtle motion cues required for fine-grained quality assessment. Third, skeleton-only representations lack environmental context, which is often important for evaluating action quality when interactions with objects or scene elements are involved.
Since quality prediction largely overlaps with video-based AQA, we focus this section on skeletal acquisition and representation learning, which most clearly reflect the unique challenges of skeleton-based AQA. Given the relatively limited number of skeleton-based studies, this section emphasizes their distinctive modeling challenges rather than providing an exhaustive methodological breakdown.

\subsubsection{Skeletal Acquisition Methods}
High-quality skeletal data significantly affects model performance. Acquisition methods include RGB cameras, depth sensors, and optical motion capture systems, each with distinct trade-offs (\cref{tab:pose-estimation}).
\textbf{RGB camera-based methods}~\cite{lei2020learning,gallardo2024gymetricpose,fieraru2021aifit,abedi2023cross,zhou2024attention} estimate 2D/3D poses from video. They are low-cost and accessible, but face challenges such as occlusion, depth ambiguity, and pose inaccuracies. Improvements like 2D-to-3D lifting~\cite{gallardo2024gymetricpose}, multi-view reconstruction~\cite{fieraru2021aifit}, and context integration have expanded their use in sports, healthcare, and fitness.
\textbf{Depth sensors} (e.g., Kinect)~\cite{bruce2021skeleton,liao2020deep,du2021assessing} directly capture 3D joints with high accuracy, making them suitable for medical applications like physiotherapy~\cite{bruce2022egcn}. However, they are more costly and sensitive to the environment. Combining RGB and depth (RGB-D) leverages both modalities for enhanced pose estimation, particularly in rehabilitation.
\textbf{Optical motion capture (MoCap) systems} (e.g., Vicon, Qualisys)~\cite{wang2022skeleton,bruce2021skeleton} provide the highest precision for skeletal tracking, but require complex, expensive setups and are mainly used in controlled labs.
RGB cameras are affordable and flexible, but less precise. Depth sensors improve 3D accuracy but have environmental and cost constraints. MoCap offers unmatched accuracy at the expense of flexibility and cost.
The choice depends on the balance between accuracy and practicality.

\subsubsection{Skeletal Representation Learning Methods}
Unlike video-based representation learning, which focuses on pixel-based features from raw images or frames, skeletal representation learning directly works with abstract body joint coordinates and their dynamic interactions over time. The complexity of joint relationships makes skeletal representation learning distinct, as it emphasizes the spatial and temporal dependencies between body parts. 
This section first reviews typical feature representation methods and then introduces robust mechanisms for enhanced AQA.

\begin{table}
    \rowcolors{2}{gray!3}{gray!12}
    \centering
    \caption{Statistics of popular skeleton acquisition techniques in AQA.}
    \resizebox{\linewidth}{!}{
    \begin{tabular}{>{\hfill}m{4.6cm}<{\centering}m{2.8cm}<{\centering}m{2.2cm}<{\centering}m{4.2cm}<{\centering}}
    \toprule
    \textbf{Data Acquisition}  & \textbf{Input} & \textbf{Output} & \textbf{Typical Works}  \\
    \midrule
    OpenPose \cite{cao2017realtime}  & Image & 2D Joints &
    \cite{lei2020learning,lei2023multi} \\
    MediaPipe \cite{lugaresi2019mediapipe} & Image & 2D Joints & \cite{kryeem2023personalized,abedi2023cross} \\
    ViTPose \cite{xu2022vitpose} & Image & 2D Joints &  \cite{gallardo2024gymetricpose} \\
    CPM \cite{wei2016convolutional} & Image & 2D Joints & \cite{wang2020towardsa} \\
    MMPose \cite{mmpose2020} & Image & 3D Joints & \cite{li2024segmentation} \\
    LTHP \cite{iskakov2019learnable} & Multi-view Images & 3D Joints & \cite{zheng2023skeleton} \\
    MotioAGF.  \cite{mehraban2024motionagformer} & 2D Joints & 3D Joints & \cite{gallardo2024gymetricpose}  \\
    PosePrior \cite{zimmermann2017learning} & 2D Joints & 3D Joints & \cite{wang2020towardsa} \\
    VideoPose3D \cite{pavllo20193d} & Video & 3D Joints & \cite{joung2023contrastive,zhou2024attention,garg2023short} \\
    MubyNet \cite{zanfir2018deep} & Image & 3D Pose \& Shape & \cite{fieraru2021aifit} \\
Depth Sensors (e.g., Kinect) & Depth & 3D Joints & \cite{wang2022skeleton,bruce2022egcn,bruce2024egcn++,capecci2019kimore,li2023graph,deb2022graph,yao2023contrastive,bruce2021skeleton,liao2020deep,du2021assessing,li2024finerehab} \\
    Marker-Based (MoCap) & Human Body & 3D joints & \cite{wang2022skeleton,bruce2021skeleton} \\
    \bottomrule
    \end{tabular}
    }
    \label{tab:pose-estimation}
\end{table}

\myPara{Basic Learning Methods.}
Skeletal representation learning in AQA has evolved from handcrafted descriptors to deep neural models. Early approaches relied on manually designed features combined with traditional classifiers, such as DCT-based descriptors or self-similarity features with SVMs~\cite{lei2020learning}, but these methods struggled to capture complex kinematic patterns and generalized poorly. With the advent of deep learning, CNNs, LSTMs, GCNs, and Transformers have been explored, among which graph convolutional networks (GCNs) have become dominant due to their natural compatibility with skeletal data~\cite{liao2020deep,bruce2021skeleton,kanade2023attention}. By representing joints as nodes and spatial–temporal connections as edges, GCNs provide a principled way to model structured body-part relations~\cite{deb2022graph,li2023graph}. The introduction of ST-GCN further enabled joint spatial–temporal modeling by integrating temporal convolution~\cite{yan2018spatial}, laying the foundation for modern skeleton-based AQA. Subsequent works build upon this paradigm with enhanced strategies to improve assessment robustness~\cite{du2021assessing,bruce2022egcn,bruce2024egcn++,yao2023contrastive}.

\myPara{Robust Learning Mechanisms.}
Beyond backbone architectures, recent advances focus on improving robustness and discriminability under noisy, sparse, and variable-length skeletal data. 
We summarize four representative mechanisms.
\emph{Hierarchical body-part modeling} exploits structured priors by decomposing the skeleton into body parts or motion hierarchies~\cite{liao2020deep,wang2022skeleton,lei2023multi,deb2022graph}. 
By modeling joints, limbs, and body parts at different granularities, these methods enhance sensitivity to localized motion quality while preserving global coordination.
\emph{Multi-input and ensemble modeling} improves robustness by integrating complementary skeletal cues. 
EGCN and its extensions~\cite{bruce2022egcn,bruce2024egcn++} combine positional and orientational features through data- and model-level fusion, enabling richer motion representations.
Related works further incorporate multi-task learning, jointly optimizing quality assessment with auxiliary objectives such as abnormality detection or reference comparison~\cite{bruce2021skeleton,li2023graph}.
\emph{Contrastive and relational learning} introduces relative reasoning into skeletal representation learning. 
By comparing test actions with reference or standard motions, contrastive objectives enhance discrimination between subtle quality differences~\cite{yao2023contrastive}, complementing absolute regression-based learning.
\emph{Self-supervised learning} addresses annotation scarcity and domain variability by encouraging consistent motion representations without explicit labels~\cite{nekoui2021enhancing,du2021assessing}. 
These approaches improve generalization by learning intrinsic motion patterns before or alongside supervised quality prediction.

\begin{figure}[!ht]
    \centering
    \includegraphics[width=\linewidth,clip,trim=0 145 0 145]{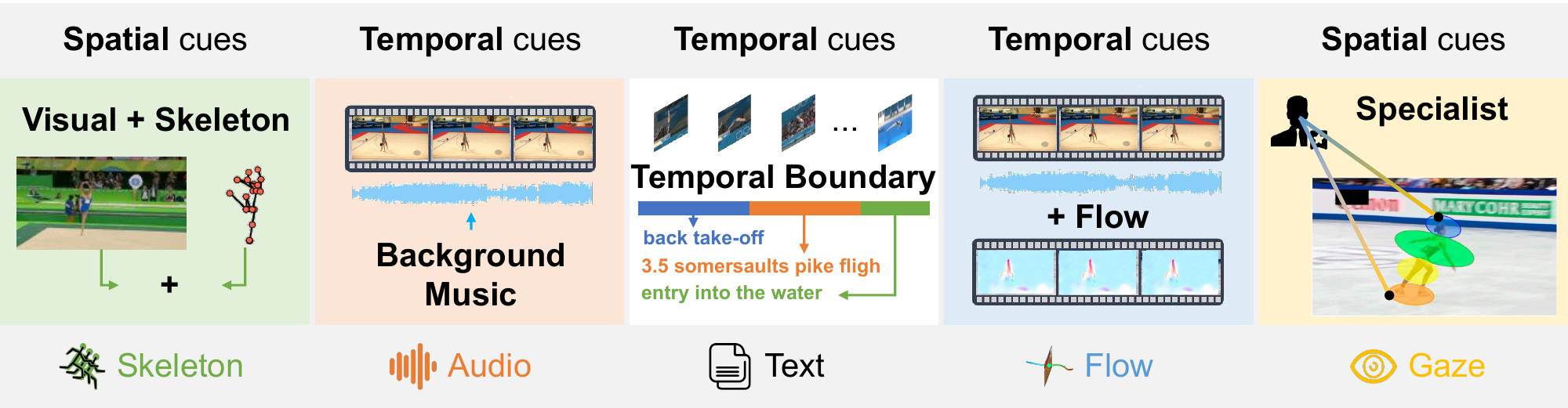}
    \caption{Illustration of typical multi-modal AQA methods.}
    \label{fig:mm-method}
    \vspace{-0.25cm}
\end{figure}

\subsection{Multi-Modal Approach}
\label{sec:taxonomy-multi_modality}
Multi-modal AQA methods improve robustness and assessment accuracy by integrating complementary information from multiple data sources, alleviating the intrinsic limitations of unimodal approaches. 
For example, RGB-based methods are sensitive to lighting and viewpoint changes, while pose or sensor signals may be noisy or incomplete. 
By fusing modalities such as audio, text, skeletons, optical flow, and wearable sensors, multi-modal AQA provides a more comprehensive understanding of action quality (see \cref{fig:mm-method}). 
Such approaches are particularly effective in domains requiring fine-grained and reliable assessment, including sports analysis \cite{xia2023skating,zhong2023contrastive,zeng2024multimodal,zhang2024narrative,xu2024vision,du2023learning,gedamu2024visual} and healthcare \cite{nagai2024mmw,hirosawa2023action,kim2024automatic,khan2022mrehab,yin2025flex}. 
We organize existing methods by auxiliary modality and discuss how each modality contributes to AQA.

\myPara{Audio-Assisted Methods}
Audio provides complementary temporal and rhythmic cues that are difficult to infer from visual motion alone, particularly in activities in which action quality is closely tied to timing, pace, or synchronization with sound. Early audio-assisted AQA studies therefore focused on rhythm-dominated domains such as figure skating~\cite{xia2023skating}, dance~\cite{zhong2023contrastive}, piano performance~\cite{parmar2021piano}, and long-term assessment~\cite{zeng2024multimodal,wang2025attentiondriven}. Skating-Mixer~\cite{xia2023skating} jointly models audio and visual streams with memory recurrent units to capture long-term temporal dependencies in skating routines, while contrastive audio–visual learning was introduced to align dance movements with musical structure and beat dynamics~\cite{zhong2023contrastive}. Subsequent works extended audio–visual fusion to richer multi-modal settings. PAMFN~\cite{zeng2024multimodal} progressively aggregates RGB, optical flow, and audio features through modality-specific and shared branches to enhance fine-grained quality estimation. To handle long-term sequences, LMAC-Net~\cite{wang2025attentiondriven} employs attention-driven fusion to align RGB, optical flow, and audio, improving temporal coherence over extended actions. More recent methods further incorporate semantic guidance: MLAVL~\cite{xu2025language} injects language prompts to align audio–visual grading with textual criteria, while MCMoE~\cite{xu2025mcmoe} uses mixture-of-experts to complete missing audio or visual modalities. These developments demonstrate that audio cues consistently enhance rhythm awareness, temporal alignment, and robustness in AQA.

\myPara{Text-Assisted Methods}
Text-assisted AQA incorporates criteria instructions and language feedback into visual assessment, enabling more context-aware parsing and interpretable evaluation. Recent works increasingly explore vision-language interaction and prompt-based assessment, reflecting a broader trend of integrating language supervision into visual reasoning. 
One line of work uses textual information to enhance visual representation learning and improve assessment performance. 
VLAKL \cite{xu2024vision} embeds domain-specific terminology into multi-level representations to address coarse action segmentation, while SGN \cite{du2023learning} adopts a teacher-student framework to transfer semantic knowledge from text to visual features.
Another line of work focuses on generating language-based feedback to improve interpretability. 
NAE-AQA \cite{zhang2024narrative} reformulates score regression as video-text matching using prompt-guided transformers, producing textual descriptions rather than numerical scores to improve readability. 
ExpertAF \cite{ashutosh2024expertaf} generates expert-like free-form feedback from RGB video and 3D pose conditioned on text prompts. 
CROSSTRAINER \cite{ashutosh2025learning} mines transferable skill attributes from video-language pairs. 
EFA \cite{qi2025explainable} introduces chain-of-thought reasoning for video-language form assessment, while QGVL \cite{xu2025quality} leverages quality-aware prompts to guide scoring.
These studies highlight the growing role of language in injecting high-level semantics and interpretability into AQA systems.

\myPara{Video-Skeleton Methods}
Video-skeleton methods combine visual appearance with explicit human kinematics, capturing both contextual cues and precise motion dynamics.
Such fusion is particularly beneficial when appearance alone is ambiguous or when fine-grained body control is critical.
Early two-stream designs integrate RGB and pose features to improve robustness \cite{li2022tai}.
Subsequent works focus on long-term and complex actions.
Zahan et al. \cite{zahan2024learning} introduce the AGF-Olympics dataset and a discriminative attention module for long sports routines.
2M-AF \cite{ding20242m} fuses RGB and skeleton streams via preference-based feature aggregation.
LucidAction \cite{dong2025lucidaction} provides multi-view RGB with 2D and 3D poses and adapts multiple AQA architectures.
Recent advances further refine pose-guided fusion.
MELO \cite{ding2025sceneaware} introduces scene-aware contrastive fusion for multi-person routines.
PGT \cite{zhang2025poseguided} injects 2D and 3D pose priors into transformers.
HP-MCoRe \cite{qi2025action} adopts hierarchical pose-guided contrastive regression.
DanceFix \cite{xu2025dancefix} demonstrates the effectiveness of RGB-pose fusion for skiing and group dance assessment.
These methods show that video-skeleton integration substantially improves fine-grained action understanding.

\myPara{Other Modality-Assisted Methods}
Beyond audio, text, and skeletons, several works incorporate additional sensing modalities to enhance AQA.
Wearable and environmental sensors provide complementary kinematic and physiological signals.
MMW-AQA \cite{nagai2024mmw} combines video, IMU, and GPS for windsurfing assessment.
Gaze information is fused with kinematics to analyze figure skating jumps \cite{hirosawa2023action}.
Optical flow is integrated with RGB using cross-attention for fine-grained evaluation \cite{kim2024automatic}.
UJ-AQA-CricketVision \cite{moodley2025i3daelstm} trains a dual-stream autoencoder over RGB and keypoints.
FLEX \cite{yin2025flex} further expands modality coverage by providing RGB, 3D pose, sEMG, and physiological signals for fitness benchmarking.
These studies illustrate the potential of sensor-rich settings for precise and robust AQA.

\subsection{Modality Selection and Trade-offs} 
\label{sec:taxonomy-guidline}
As summarized in \cref{tab:modality_guidelines}, the choice of modality in AQA reflects a fundamental trade-off between information richness, annotation cost, robustness, and computational efficiency.
Video-based methods provide the most comprehensive visual cues and remain the dominant choice for complex actions, but they are sensitive to background variations and incur high computation and annotation cost. 
Skeleton-based approaches abstract actions into joint trajectories, offering efficiency and interpretability for human-centric scenarios, yet they are vulnerable to pose estimation errors and often lack contextual cues needed for fine-grained assessment. 
Multi-modal methods further enhance robustness by integrating complementary signals, but their reliance on synchronized data acquisition, modality alignment, and full-modality availability at inference significantly limits scalability and real-world deployment.

\begin{table*}[!ht]
\centering
\rowcolors{2}{gray!3}{gray!12}
\caption{Comparison of video-based, skeleton-based, and multi-modal AQA approaches from the perspective of supervision, robustness, efficiency, and application scenarios.}
\label{tab:modality_guidelines}
\resizebox{\linewidth}{!}{
\begin{tabular}{lcccccc}
\toprule
\textbf{Modality} &
\textbf{Information Richness} &
\textbf{Annotation Cost} &
\textbf{Robustness} &
\textbf{Computational Cost} &
\textbf{Interpretability} &
\textbf{Suitable Scenarios} \\
\midrule
Video-based (RGB) 
& High 
& High 
& Medium 
& High 
& Low 
& Sports judging, in-the-wild videos \\

Skeleton-based 
& Medium 
& Medium 
& Medium--Low 
& Low 
& High 
& Rehabilitation, technique analysis \\

Multi-modal 
& Very High 
& Very High 
& High 
& Very High 
& Medium--High 
& Medical assessment, expert analysis \\
\bottomrule
\end{tabular}
}
\end{table*}

In practice, modality selection should be guided by application constraints rather than performance alone. 
Skeleton-based methods are preferable when precise kinematic analysis and efficiency are prioritized, video-based approaches suit scenarios where visual context is essential, and multi-modal designs are most appropriate when annotation cost and inference complexity are acceptable in exchange for maximum robustness. 
An emerging direction is to design modality-flexible frameworks that can selectively activate auxiliary signals, balancing performance and practicality.

\section{Task-Specific Applications} \label{sec:app}

Despite strong benchmark performance, AQA still faces real-world challenges in label scarcity, continual adaptation, and interpretability. To address these challenges, we identify several underexplored research directions for practical AQA deployment. 
\myPara{Semi-Supervised AQA} \label{sec:challenges_labeled_data}
\textit{Labeled data scarcity} remains a fundamental bottleneck for AQA, as high-quality annotations often require expert knowledge and are costly to obtain. 
Semi-supervised AQA methods~\cite{yun2024semi,zhang2022semi,gedamu2024self,ye2025decoupling} aim to alleviate this issue by exploiting large amounts of unlabeled data together with limited labeled samples, using techniques such as self-training, consistency regularization, and pseudo-labeling.
Early work such as S$^4$AQA~\cite{zhang2022semi} explored masked segment recovery to learn robust representations from unlabeled videos, while TRS-AQA~\cite{yun2024semi} adopted a teacher–student paradigm to generate reliable pseudo labels.
These studies indicate that semi-supervised learning is a promising direction for scaling AQA to new domains, though ensuring pseudo-label reliability remains an open challenge.

\myPara{Continual AQA} \label{sec:challenges_variations}
\textit{Non-stationary variations} in action execution, recording conditions, and user populations pose significant challenges to the robustness of AQA systems.
Continual AQA~\cite{li2024continual,zhou2024magr} investigates how models can incrementally adapt to new data distributions without catastrophic forgetting, drawing on advances in continual learning~\cite{wang2024comprehensive}.
Representative methods include PECoP~\cite{dadashzadeh2024pecop}, which mitigates domain shift during incremental training, methods that explicitly address the introduction of new action categories~\cite{li2024continual}, and MAGR~\cite{zhou2024magr,zhou2025continual}, which leverages manifold projection and graph regularization to stabilize learning while preserving privacy.
While these methods show encouraging progress, balancing adaptability, memory efficiency, and long-term stability remains an open research problem for practical AQA systems.

\myPara{Interpretable AQA} \label{sec:challenges_feedback}
\textit{Limited interpretability} is also a major barrier to the real-world adoption of AQA, as numerical scores alone provide little actionable guidance to users.
Interpretable AQA methods aim to augment predictions with understandable feedback.
Narrative feedback approaches~\cite{fieraru2021aifit,okamoto2024hierarchical,zhang2024narrative,li2024techcoach,ashutosh2024expertaf} usually generate descriptive explanations, error diagnoses, or coaching-style comments through vision-language foundation models.
Systems such as AIFit~\cite{fieraru2021aifit}, NSAQA~\cite{okamoto2024hierarchical}, NAE-AQA~\cite{zhang2024narrative}, ExpertAF~\cite{ashutosh2024expertaf}, and TechCoach~\cite{li2024techcoach} illustrate different levels of interpretability, ranging from keypoint-level analysis to free-form textual feedback.
Complementary visual feedback mechanisms~\cite{fieraru2021aifit,zhou2023video,ashutosh2024expertaf}, including motion overlays and side-by-side comparisons, further help users understand performance differences.
Despite this progress, generating informative explanations remains an open challenge.

\begin{table*}
    \centering
    \caption{
    Overview of popular AQA datasets, including modality, domain, number of classes, sample size, average frames, annotations, and access URLs. Datasets span sports, skill assessment, and healthcare. 
    }
    \label{tab:dataset}
    \setlength{\tabcolsep}{5pt}
    \resizebox{\linewidth}{!}{
    \begin{NiceTabular}{rcccccccc}[colortbl-like]
    \toprule
    \textbf{Dataset} & \textbf{Year} & \textbf{Modality}& \textbf{Domain} & \textbf{\# Class} & \textbf{\# Samples} & \textbf{\# Average Frames} & \textbf{Annotations} & \textbf{URL}  \\
    \midrule
    \rowcolor[gray]{0.92} MIT Olympic \cite{pirsiavash2014assessing} & 2014 &
    \makecell[c]{Video \\ 2D Skeleton} &
    Sports & 2 & 309 & 
    \makecell[c]{Dive: 150 \\ Figure Skate: 4200} &
    Score &
    \href{https://userpages.cs.umbc.edu/hpirsiav/quality.html}{\faExternalLink}
    \\
    \rowcolor[gray]{0.98} UNLV Olympic \cite{parmar2017learning} & 2017 &
    Video &
    Sports & 3 & 717 &
    \makecell[c]{Dive: 150 \\ Figure Skate: 4200 \\ Vault: 75} &
    Score &
    \href{http://rtis.oit.unlv.edu/datasets.html}{\faExternalLink}
    \\
    \rowcolor[gray]{0.92} AQA-7 \cite{parmar2019action} & 2019 &
    Video &
    Sports & 7 & 1189 &
    \makecell[c]{Dive: 97 105 156 \\ Vault: 87 \\ Big Air: 122 132 \\ Trampoline: 634} &
    Score &
    \href{http://rtis.oit.unlv.edu/datasets/}{\faExternalLink}
    \\
    \rowcolor[gray]{0.98} MTL-AQA \cite{parmar2019and} & 2019 &
    Video &
    Sports & 16 & 1412 & 
    96 &
    Score &
    \href{https://github.com/Luciferbobo/DAE-AQA}{\faExternalLink}
    \\
    \rowcolor[gray]{0.92} Fis-V \cite{xu2019learning} & 2019 &
    Video &
    Sports & 1 & 500 &
    4300 &
    TES PCS &
    \href{https://github.com/chmxu/MS_LSTM}{\faExternalLink}
    \\
    \rowcolor[gray]{0.98} RG \cite{zeng2020hybrid} & 2020 &
    Video &
    Sports & 1 & 250 &
    2375 &
    \makecell[c]{Difficulty \\ Execution \\ Total} &
    \href{https://github.com/qinghuannn/ACTION-NET}{\faExternalLink}
    \\
    \rowcolor[gray]{0.92} FineDiving \cite{xu2022finediving} & 2022 &
    Video &
    Sports & 52 & 3000 &
    105 & 
    \makecell[c]{Step \\ Score} &
    \href{https://github.com/xujinglin/FineDiving}{\faExternalLink}
    \\
    \rowcolor[gray]{0.98} TaiChi-24 \cite{li2022tai} & 2022 &
    \makecell[c]{RGB-D \\ 3D Skeleton} &
    Sports & 24 & 1408 &
    72-695 &
    Score &
    -- 
    \\
    \rowcolor[gray]{0.92} FS1000 \cite{xia2023skating} & 2023 &
    \makecell[c]{Video \\ Audio} &
    Sports & 7 & 1604 &
    5000 & 
    \makecell[c]{TES PCS \\ Detailed PCS} &
    \href{https://github.com/AndyFrancesco29/Audio-Visual-Figure-Skating}{\faExternalLink}
    \\
    \rowcolor[gray]{0.98} FineFS \cite{ji2023localization} & 2023 &
    \makecell[c]{Video \\ 2D/3D Skeleton} &
    Sports & 4 & 1167 &
    5000 &
    \makecell[c]{Detailed Score \\ Subaction Class \\ Segmentation} &
    \href{https://github.com/yanliji/FineFS-dataset}{\faExternalLink}
    \\
    \rowcolor[gray]{0.92} LOGO \cite{zhang2023logo} & 2023 &
    Video &
    Sports & 12 & 200 &
    5100 &
    \makecell[c]{Action Class \\ Formation \\ Score} &
    \href{https://github.com/shiyi-zh0408/LOGO}{\faExternalLink}
    \\ 
    \rowcolor[gray]{0.98} LucidAction \cite{dong2025lucidaction} & 2024 &
    \makecell[c]{Video \\ 2D/3D Skeleton} &
    Sports & 259 & 6702 &
    103 &
    \makecell[c]{Score \\ Penalty} &
    -- 
    \\
    \rowcolor[gray]{0.92} FineDiving-HM \cite{xu2024fineparser} & 2024 &
    \makecell[c]{Video \\ Mask} &
    Sports & 52 & 3000 &
    105 &
    \makecell[c]{Mask + Score} &
    \href{https://github.com/PKU-ICST-MIPL/FineParser_CVPR2024}{\faExternalLink}
    \\
    \rowcolor[gray]{0.98} Trampoline-AQA \cite{lin2025enhancing} & 2025 &
    \makecell[c]{Video \\ Optical Flow} &
    Sports & 1 & 206 &
    \makecell[c]{\(\approx\)902 \\ (30.05 s @30fps)} &
    \makecell[c]{Score Label \\ Competition Label \\ Meta Label} &
    \href{https://zenodo.org/records/16195090}{\faExternalLink}
    \\
    \rowcolor[gray]{0.92} UJ-AQA-CricketVision \cite{moodley2025i3daelstm} & 2025 &
    \makecell[c]{Video \\ 2D Skeleton} &
    Sports & 1 & 8500 &
    28 &
    \makecell[c]{Phase Scores} &
    \href{https://github.com/dvanderhaar/uj-aqa-cricketvision}{\faExternalLink}
    \\
    \rowcolor[gray]{0.98} FineDiving-Pose \cite{qi2025action} & 2025 &
    \makecell[c]{Video \\ 2D/3D Skeleton} &
    Sports & 1 & 3,000 &
    96 &
    \makecell[c]{Pose + Score} &
    \href{https://github.com/Lumos0507/HP-MCoRe}{\faExternalLink}
    \\
    \rowcolor[gray]{0.92} BDJ \cite{wang2025adaptive} & 2025 &
    Video &
    Sports & 1 & 920 &
    400 &
    \makecell[c]{Action Specification Score \\ Execution Level Score  \\ Total Score} &
    \href{https://github.com/han3364/MAN}{\faExternalLink}
    \\
    \rowcolor[gray]{0.98} FLEX \cite{yin2025flex} & 2025 &
    \makecell[c]{Video \\ 3D Pose \\ sEMG \\ Physiological} &
    Sports & 20 & 7500 &
    234 &
    \makecell[c]{Skill Score \\ Knowledge Graph} &
    \href{https://haoyin116.github.io/FLEX_Dataset}{\faExternalLink}
    \\
    \rowcolor[gray]{0.92} CoT-AFA \cite{qi2025explainable} & 2025 &
    \makecell[c]{Video \\ Text} &
    Sports & 141 & 3,392 &
    108 &
    \makecell[c]{Form + CoT Labels} &
    \href{https://github.com/MICLAB-BUPT/EFA}{\faExternalLink}
    \\

\rowcolor[gray]{0.98} JIGSAWS \cite{gao2014jhu} & 2014 &
    Video &
    Skill Assessment & 3 & 103 &
    -- &
    \makecell[c]{Surgemes Class \\ Rating} &
    \href{https://cirl.lcsr.jhu.edu/research/hmm/datasets/jigsaws_release/}{\faExternalLink}
    \\
    \rowcolor[gray]{0.92} EPIC-Skills \cite{doughty2018s} & 2018 &
    Video &
    Skill Assessment & 7 & 216 &
    -- &
    Relative Rank &
    \href{https://github.com/hazeld/rank-aware-attention-network}{\faExternalLink}
    \\
    \rowcolor[gray]{0.98} BEST \cite{doughty2019pros} & 2019 &
    Video &
    Skill Assessment & 5 & 500 &
    6400 &
    Relative Rank &
    \href{https://github.com/hazeld/rank-aware-attention-network}{\faExternalLink}
    \\
    \rowcolor[gray]{0.92} PISA \cite{parmar2021piano} & 2021 &
    \makecell[c]{Video \\ Audio} &
    Skill Assessment & 1 & 992 &
    160 &
    \makecell[c]{Skill Level \\ Difficulty} &
    \href{https://github.com/ParitoshParmar/Piano-Skills-Assessment}{\faExternalLink}
    \\
    \rowcolor[gray]{0.98} TAQR \cite{fang2024better} & 2024 &
    Video &
    Skill Assessment & 4 & 300 &
    488 &
    \makecell[c]{Relative Rank} &
    \href{https://github.com/MingZier/TAQR-Dataset}{\faExternalLink}
    \\
    \rowcolor[gray]{0.92} EgoExo-Fitness \cite{li2024egoexo} & 2024 &
    \makecell[c]{Video (Ego/Exo)} &
    Skill Assessment & 12 & 6,131 &
    \makecell[c]{600 \\ 10-30 s @ 30fps} &
    \makecell[c]{Score \\ Boundaries \\ Comment} &
    \href{https://github.com/iSEE-Laboratory/EgoExo-Fitness}{\faExternalLink}
    \\
    \rowcolor[gray]{0.98} EgoExoLearn \cite{huang2024egoexolearn} & 2024 &
    Video &
    Skill Assessment & 8 & 3304 &
    250  &
    \makecell[c]{Relative Rank} &
    \href{https://github.com/OpenGVLab/EgoExoLearn}{\faExternalLink}
    \\ 
    \rowcolor[gray]{0.92} BASKET \cite{pan2025basket} & 2025 &
    Video &
    Skill Assessment & 21 & 66,000 &
    \makecell[c]{16,200\\ ($\sim$9 min @ 30fps)}  &
    Score &
    \href{https://github.com/yulupan00/BASKET}{\faExternalLink}
    \\
    \rowcolor[gray]{0.98} TAQA \cite{fang2026a} & 2026 &
    \makecell[c]{Video \\ Text} &
    Skill Assessment & 4 & 3738 &
    348 &
    \makecell[c]{Score \\ Text Prompt} &
    \href{https://github.com/MingZier/TAQA-Dataset}{\faExternalLink}
    \\
\rowcolor[gray]{0.92} UI-PRMD \cite{vakanski2018data} & 2018 &
    \makecell[c]{3D Skeleton \\ Joint Pos. \& Ori.} &
    Healthcare & 10 & 1326 &
    -- &
    Binary Class &
    \href{https://webpages.uidaho.edu/ui-prmd/}{\faExternalLink}
    \\
    \rowcolor[gray]{0.98} KIMORE \cite{capecci2019kimore} & 2019 &
    \makecell[c]{Video, 3D Skeleton \\ Joint Pos. \& Ori.} &
    Healthcare & 5 & 1560 &
    -- &
    Score &
    \href{https://vrai.dii.univpm.it/content/kimore-dataset}{\faExternalLink}
    \\
    \rowcolor[gray]{0.92} EHE \cite{bruce2021skeleton} & 2021 &
    \makecell[c]{3D Skeleton \\ Joint Pos. \& Ori.} &
    Healthcare & 6 & 869 &
    -- &
    Binary Class &
    \href{https://github.com/bruceyo/egcnplusplus/tree/main/EHE_dataset}{\faExternalLink}
    \\
    \rowcolor[gray]{0.98} FineRehab \cite{li2024finerehab} & 2024 &
    \makecell[c]{Video, 3D Skeleton \\ Joint Pos. \& Ori.} &
    Healthcare & 16 & 4215 &
    -- &
    Score &
    \href{https://bsu3dvlab.github.io/FineRehab/}{\faExternalLink}
    \\
    \rowcolor[gray]{0.92} GAIA \cite{chen2024gaia} & 2024 &
    Video &
    AIGC & 510 & 9180 &
    70 &
    \makecell[c]{Subject \\ Completeness \\ Interaction} &
    \href{https://github.com/zijianchen98/GAIA}{\faExternalLink}
    \\ 
    \rowcolor[gray]{0.98} Human-AGVQA \cite{zhang2024human} & 2025 &
    \makecell[c]{Video \\ Text} &
    AIGC & 44 & 6000 &
    32 &
    \makecell[c]{Quality Ratings} &
    \href{https://github.com/zczhang-sjtu/GHVQ.git}{\faExternalLink}
    \\
    \bottomrule
    \end{NiceTabular}
    }
\end{table*}

\section{Dataset and Benchmark} \label{sec:db}
We review representative datasets and observe that video-based AQA is the most common setting. Thus, we use it as a case study to build a comprehensive benchmark. 

\subsection{Datasets} \label{sec:db_dataset}
As shown in \cref{tab:dataset}, existing AQA datasets exhibit three evolving trends. 

First, there is a growing emphasis on \textbf{modality diversity and fine-grained annotation}, moving beyond single-stream RGB videos toward richer supervision signals. 
Early datasets such as MIT Olympic~\cite{pirsiavash2014assessing}, UNLV Olympic~\cite{parmar2017learning}, and AQA-7~\cite{parmar2019action} mainly provide video-level scores, whereas more recent benchmarks introduce multi-modal inputs and structured labels, including step-level annotations in FineDiving~\cite{xu2022finediving}, detailed subaction scores and segmentation in FineFS~\cite{ji2023localization}, and pose-enhanced supervision in FineDiving-Pose~\cite{qi2025action}. 
Healthcare-oriented datasets such as FineRehab~\cite{li2024finerehab} and KIMORE~\cite{capecci2019kimore} further incorporate 3D skeletal signals to support precise AQA.

Second, datasets are rapidly expanding in \textbf{scale and temporal coverage}, enabling more robust training and evaluation of data-hungry models. 
While earlier benchmarks are limited to a few hundred samples, large-scale datasets such as FS1000~\cite{xia2023skating}, BASKET~\cite{pan2025basket}, and GAIA~\cite{chen2024gaia} provide thousands to tens of thousands of videos, often with long temporal durations. 
Recent long-sequence datasets, including LOGO~\cite{zhang2023logo} and Trampoline-AQA~\cite{lin2025enhancing}, explicitly target minute-level actions, posing new challenges for temporal modeling and efficiency.

Third, AQA datasets are expanding into \textbf{broader application domains}, reflecting the practical diversification of AQA research. 
AQA datasets have steadily expanded into more diverse application domains: early extensions focused on skill assessment and healthcare (e.g., EPIC-Skills~\cite{doughty2018s}, UI-PRMD~\cite{vakanski2018data}, EHE~\cite{bruce2021skeleton}), while more recent releases further broaden the scope to teaching assessment (e.g., TAQR~\cite{fang2024better}, TAQA~\cite{fang2026a}) and to emerging or fine-grained settings such as AI-generated action evaluation (GAIA~\cite{chen2024gaia}, Human-AGVQA~\cite{zhang2024human}) and traditional activities with specialized criteria (e.g., BDJ~\cite{wang2025adaptive}).

\subsection{A Unified AQA Benchmark} \label{sec:db_benchmark}
Existing AQA studies are often evaluated on disparate datasets with inconsistent protocols, hindering fair comparisons and obscuring real progress across methods. 
To mitigate this issue, we construct a unified benchmark centered on \emph{video-based AQA}, which remains the most mature, widely adopted, and reproducible research setting in the literature. 
Skeleton-based and multi-modal approaches are not included at this stage, primarily due to the limited availability of standardized datasets and open-source implementations that support controlled benchmarking. 

\subsubsection{Experimental Setting}
Our benchmark covers six widely used datasets, seven state-of-the-art baseline methods, and seven evaluation metrics.

\myPara{Benchmark Datasets}
We select datasets according to three criteria: 
(1) public availability with clearly defined train/test splits, 
(2) coverage of both short-term and long-term actions, and 
(3) compatibility with standard video-based pipelines without requiring additional sensors or annotations.
Accordingly, we include short-term benchmarks such as MTL-AQA~\cite{parmar2019and}, AQA-7~\cite{parmar2019action}, and FineDiving~\cite{xu2022finediving}, as well as long-term datasets including Fis-V~\cite{xu2019learning}, RG~\cite{zeng2020hybrid}, and LOGO~\cite{zhang2023logo}.  

\myPara{Benchmark Methods}
We include representative baselines selected based on four criteria: 
(1) publicly available implementations to ensure reproducibility; 
(2) competitive performance reported on standard AQA benchmarks; 
(3) methodological diversity, covering major paradigms in \cref{sec:taxonomy}; and 
(4) compatibility with unified evaluation settings (see \cref{sec_unified_framework-evaluation_metrics}) for fair comparison. Specifically, USDL~\cite{tang2020uncertainty} and DAE~\cite{zhang2024auto} represent uncertainty-aware direct regression, 
CoRe~\cite{yu2021group} and T$^2$CR~\cite{ke2024two} represent contrastive regression and multi-view reasoning, 
GDLT~\cite{xu2022likert} and CoFInAl~\cite{zhou2024cofinal} represent transformer-based methods, 
and HGCN~\cite{zhou2023hierarchical} represents structured graph reasoning.

\begin{table*}
    \newcommand{\stdplaceholder}[1]{\textcolor{gray}{\scriptsize #1}}
    \centering
    \caption{Results of our video-based AQA benchmark. The training time unit is in hours. The average SRCC is calculated using Fisher-z transformation to ensure comparability across datasets. The best results are presented in \textbf{bold}, while the second-best results are \underline{underlined}.}
    \setlength{\tabcolsep}{3pt}
    \resizebox{\linewidth}{!}{
    \begin{tabular}{rrcccccccccccccccccccccccc }
    \toprule
    \multirow{2.5}{*}{\textbf{Method}} & \multirow{2.5}{*}{\textbf{Publisher}} & \multicolumn{4}{c}{\textbf{MTL-AQA} \cite{parmar2019and}} & \multicolumn{4}{c}{\textbf{AQA-7} $\color{gray}^\text{Average}$ \cite{parmar2019action}} & \multicolumn{4}{c}{\textbf{FineDiving} \cite{xu2022finediving}} & \multicolumn{4}{c}{\textbf{RG} $\color{gray}^\text{Average}$ \cite{zeng2020hybrid}} & \multicolumn{4}{c}{\textbf{Fis-V} $\color{gray}^\text{Average}$ \cite{xu2019learning}} & \multicolumn{4}{c}{\textbf{LOGO} \cite{zhang2023logo}} \\ 
    \cmidrule(lr){3-6} \cmidrule(lr){7-10} \cmidrule(lr){11-14} \cmidrule(lr){15-18} \cmidrule(lr){19-22} \cmidrule(lr){23-26}
    & & SRCC & MSE & rMSE & Time & SRCC & MSE & rMSE & Time & SRCC & MSE & rMSE & Time & SRCC & MSE & rMSE & Time & SRCC & MSE & rMSE & Time & SRCC & MSE & rMSE & Time \\
    \midrule
    \rowcolor[gray]{0.92} MUSDL \cite{tang2020uncertainty} & CVPR'20 & 0.9350 & 39.7753 & 0.3642 & 11.32 & \underline{0.8292} & 268.1272 & 3.0767 & \underline{1.57} & 0.8812 & 53.6996 & 0.4927 & 14.08 & 0.4897 & 12.0564 & 4.3557 & \textbf{0.01} & 0.3514 & 478.8537 & 41.1493 & 0.07 & 0.7044 & 368.2800 & \underline{3.6828} & \textbf{0.02} \\
    \rowcolor[gray]{0.98}  &  & \stdplaceholder{0.0052} & \stdplaceholder{5.9591} & \stdplaceholder{0.0545} & \stdplaceholder{0.03} & \stdplaceholder{0.0126} & \stdplaceholder{10.1333} & \stdplaceholder{0.1133} & \stdplaceholder{0.04} & \stdplaceholder{0.0017} & \stdplaceholder{1.4217} & \stdplaceholder{0.0130} & \stdplaceholder{0.03} & \stdplaceholder{0.0437} & \stdplaceholder{0.9665} & \stdplaceholder{0.3443} & \stdplaceholder{0.00} & \stdplaceholder{0.0112} & \stdplaceholder{0.3705} & \stdplaceholder{0.0592} & \stdplaceholder{0.00} & \stdplaceholder{0.0199} & \stdplaceholder{1.2255} & \stdplaceholder{0.1291} & \stdplaceholder{0.01} \\
    \rowcolor[gray]{0.92} CoRe \cite{yu2021group} & ICCV'21 & 0.9519 & 32.9234 & 0.3015 & 46.81 & 0.7838 & 256.4000 & 2.8641 & 4.44 & \underline{0.9413} & \underline{24.4000} & \underline{0.2440} & 26.62 & 0.7348 & \textbf{6.0166} & \textbf{2.2178} & 0.12 & 0.7253 & \underline{183.9800} & \underline{1.8398} & 0.20 & 0.5852 & 45.3446 & 4.7783 & 0.10 \\
    \rowcolor[gray]{0.98}  &  & \stdplaceholder{0.0311} & \stdplaceholder{18.6081} & \stdplaceholder{0.1704} & \stdplaceholder{0.90} & \stdplaceholder{0.0287} & \stdplaceholder{22.9929} & \stdplaceholder{0.2513} & \stdplaceholder{0.02} & \stdplaceholder{0.0647} & \stdplaceholder{25.7334} & \stdplaceholder{0.2361} & \stdplaceholder{1.72} & \stdplaceholder{0.0202} & \stdplaceholder{0.4138} & \stdplaceholder{0.1466} & \stdplaceholder{0.01} & \stdplaceholder{0.0082} & \stdplaceholder{0.5675} & \stdplaceholder{0.0751} & \stdplaceholder{0.01} & \stdplaceholder{0.0229} & \stdplaceholder{2.6094} & \stdplaceholder{0.2750} & \stdplaceholder{0.01} \\
    \rowcolor[gray]{0.92} GDLT \cite{xu2022likert} & CVPR'22 & 0.9395 & 43.5769 & 0.3990 & \textbf{10.84} & 0.8057 & \underline{238.9922} & \textbf{2.7120} & 1.73 & 0.9342 & 27.9100 & 0.2791 & 8.84 & \textbf{0.7584} & \underline{6.2177} & \underline{2.2767} & 0.04 & 0.7203 & 190.2050 & 1.9021 & 0.04 & \textbf{0.7453} & 401.5500 & 4.0155 & 0.04 \\
    \rowcolor[gray]{0.98}  &  & \stdplaceholder{0.0019} & \stdplaceholder{1.3744} & \stdplaceholder{0.0126} & \stdplaceholder{0.05} & \stdplaceholder{0.0199} & \stdplaceholder{28.5499} & \stdplaceholder{0.3179} & \stdplaceholder{0.01} & \stdplaceholder{0.0038} & \stdplaceholder{1.7906} & \stdplaceholder{0.0164} & \stdplaceholder{0.08} & \stdplaceholder{0.0163} & \stdplaceholder{0.4778} & \stdplaceholder{0.1721} & \stdplaceholder{0.00} & \stdplaceholder{0.0182} & \stdplaceholder{0.9873} & \stdplaceholder{0.1302} & \stdplaceholder{0.00} & \stdplaceholder{0.0994} & \stdplaceholder{7.6091} & \stdplaceholder{0.8018} & \stdplaceholder{0.00} \\
    \rowcolor[gray]{0.92} HGCN \cite{zhou2023hierarchical} & TCSVT'23 & \underline{0.9522} & \underline{30.7432} & \underline{0.2815} & 14.79 & \textbf{0.8451} & \textbf{235.1635} & \underline{2.7642} & \textbf{1.53} & 0.9383 & 24.7700 & 0.2477 & 9.14 & 0.7121 & 6.8748 & 2.5240 & \underline{0.03} & \underline{0.7270} & 227.8600 & 2.2786 & \textbf{0.03} & \underline{0.7082} & 369.0600 & 3.6906 & 0.04 \\
    \rowcolor[gray]{0.98}  &  & \stdplaceholder{0.0023} & \stdplaceholder{2.6689} & \stdplaceholder{0.0244} & \stdplaceholder{0.06} & \stdplaceholder{0.0121} & \stdplaceholder{8.9000} & \stdplaceholder{0.0942} & \stdplaceholder{0.04} & \stdplaceholder{0.0018} & \stdplaceholder{4.1503} & \stdplaceholder{0.0381} & \stdplaceholder{0.06} & \stdplaceholder{0.0392} & \stdplaceholder{1.2362} & \stdplaceholder{0.4725} & \stdplaceholder{0.00} & \stdplaceholder{0.0140} & \stdplaceholder{2.7601} & \stdplaceholder{0.2424} & \stdplaceholder{0.00} & \stdplaceholder{0.0492} & \stdplaceholder{9.2508} & \stdplaceholder{0.9748} & \stdplaceholder{0.01} \\
    \rowcolor[gray]{0.92} DAE \cite{zhang2024auto} & NCAA'24 & 0.9497 & 31.3355 & 0.2869 & \underline{11.11} & 0.7916 & 307.4535 & 3.4763 & 1.99 & 0.9350 & 26.5100 & 0.2651 & \textbf{8.58} & 0.7412 & 7.1299 & 2.7019 & 0.03 & \textbf{0.7447} & \textbf{181.1600} & \textbf{1.8116} & \underline{0.03} & 0.6701 & 356.6300 & \textbf{3.5663} & \underline{0.02} \\
    \rowcolor[gray]{0.98}  &  & \stdplaceholder{0.0009} & \stdplaceholder{2.9995} & \stdplaceholder{0.0275} & \stdplaceholder{0.02} & \stdplaceholder{0.0136} & \stdplaceholder{32.3747} & \stdplaceholder{0.4067} & \stdplaceholder{0.01} & \stdplaceholder{0.0009} & \stdplaceholder{1.3459} & \stdplaceholder{0.0123} & \stdplaceholder{0.06} & \stdplaceholder{0.0032} & \stdplaceholder{0.5730} & \stdplaceholder{0.2287} & \stdplaceholder{0.00} & \stdplaceholder{0.0058} & \stdplaceholder{2.1296} & \stdplaceholder{0.2655} & \stdplaceholder{0.00} & \stdplaceholder{0.0196} & \stdplaceholder{7.5021} & \stdplaceholder{0.7906} & \stdplaceholder{0.01} \\
    \rowcolor[gray]{0.92} T$^2$CR \cite{ke2024two} & INFS‘24 & \textbf{0.9529} & \textbf{29.8640} & \textbf{0.2735} & 49.62 & 0.7910 & 303.2644 & 3.4754 & 4.98 & \textbf{0.9424} & \textbf{23.5100} & \textbf{0.2351} & 30.56 & 0.6581 & 7.0013 & 2.5215 & 0.28 & 0.6898 & 236.1867 & 2.3619 & 1.09 & 0.5199 & \underline{43.3161} & 4.5645 & 0.26 \\
    \rowcolor[gray]{0.98}  &  & \stdplaceholder{0.0012} & \stdplaceholder{0.1800} & \stdplaceholder{0.0020} & \stdplaceholder{0.52} & \stdplaceholder{0.0141} & \stdplaceholder{11.3500} & \stdplaceholder{0.1300} & \stdplaceholder{0.07} & \stdplaceholder{0.0012} & \stdplaceholder{0.1612} & \stdplaceholder{0.0015} & \stdplaceholder{0.03} & \stdplaceholder{0.0172} & \stdplaceholder{0.4346} & \stdplaceholder{0.1602} & \stdplaceholder{0.02} & \stdplaceholder{0.0131} & \stdplaceholder{0.7144} & \stdplaceholder{0.0834} & \stdplaceholder{0.06} & \stdplaceholder{0.0767} & \stdplaceholder{3.8466} & \stdplaceholder{0.4053} & \stdplaceholder{0.04} \\
    \rowcolor[gray]{0.92} CoFInAl \cite{zhou2024cofinal} & IJCAI'24 & 0.9461 & 37.7907 & 0.3461 & 14.80 & 0.8195 & 249.9134 & 2.7769 & 1.57 & 0.9317 & 36.4681 & 0.2887 & \underline{8.64} & \underline{0.7534} & 10.8178 & 4.0537 & 0.08 & 0.6974 & 401.9300 & 4.0193 & 0.05 & 0.5972 & \textbf{38.6218} & 4.0698 & 0.05 \\
    \rowcolor[gray]{0.98}  &  & \stdplaceholder{0.0005} & \stdplaceholder{1.0202} & \stdplaceholder{0.0094} & \stdplaceholder{0.05} & \stdplaceholder{0.0127} & \stdplaceholder{9.2500} & \stdplaceholder{0.1000} & \stdplaceholder{0.04} & \stdplaceholder{0.0017} & \stdplaceholder{0.5107} & \stdplaceholder{0.0120} & \stdplaceholder{0.81} & \stdplaceholder{0.0129} & \stdplaceholder{1.0906} & \stdplaceholder{0.3848} & \stdplaceholder{0.01} & \stdplaceholder{0.0151} & \stdplaceholder{5.5130} & \stdplaceholder{0.5727} & \stdplaceholder{0.01} & \stdplaceholder{0.0777} & \stdplaceholder{1.4044} & \stdplaceholder{0.1480} & \stdplaceholder{0.00} \\
    \midrule
    \multirow{2.5}{*}{\textbf{Method}} & \multirow{2.5}{*}{\textbf{Publisher}} & \multicolumn{4}{c}{\textbf{Diving} $\color{gray}^\text{AQA-7}$ \cite{parmar2019action}} & \multicolumn{4}{c}{\textbf{Gym Vault} $\color{gray}^\text{AQA-7}$ \cite{parmar2019action}} & \multicolumn{4}{c}{\textbf{BigSki.} $\color{gray}^\text{AQA-7}$ \cite{parmar2019action}} & \multicolumn{4}{c}{\textbf{BigSnow.} $\color{gray}^\text{AQA-7}$ \cite{parmar2019action}} & \multicolumn{4}{c}{\textbf{Sync. 3m} $\color{gray}^\text{AQA-7}$ \cite{parmar2019action}} & \multicolumn{4}{c}{\textbf{Sync. 10m} $\color{gray}^\text{AQA-7}$ \cite{parmar2019action}} \\ 
    \cmidrule(lr){3-6} \cmidrule(lr){7-10} \cmidrule(lr){11-14} \cmidrule(lr){15-18} \cmidrule(lr){19-22} \cmidrule(lr){23-26}
    & & SRCC & MSE & rMSE & Time & SRCC & MSE & rMSE & Time & SRCC & MSE & rMSE & Time & SRCC & MSE & rMSE & Time & SRCC & MSE & rMSE & Time & SRCC & MSE & rMSE & Time \\
    \midrule
    \rowcolor[gray]{0.92} MUSDL \cite{tang2020uncertainty} & CVPR'20 & \underline{0.8738} & 129.5963 & 1.2960 & 3.21 & 0.7300 & 133.8101 & 2.1927 & 1.44 & 0.5369 & 729.6780 & 8.4625 & \underline{1.44} & \textbf{0.7109} & \textbf{366.3390} & \textbf{3.6634} & \underline{1.64} & \underline{0.9205} & 126.1327 & 1.5508 & \textbf{0.85} & \underline{0.9416} & \textbf{124.1512} & \textbf{1.3039} & \underline{0.86} \\
    \rowcolor[gray]{0.98}  &  & \stdplaceholder{0.0120} & \stdplaceholder{6.5000} & \stdplaceholder{0.0650} & \stdplaceholder{0.04} & \stdplaceholder{0.0115} & \stdplaceholder{5.8000} & \stdplaceholder{0.0950} & \stdplaceholder{0.04} & \stdplaceholder{0.0180} & \stdplaceholder{25.0000} & \stdplaceholder{0.2800} & \stdplaceholder{0.05} & \stdplaceholder{0.0160} & \stdplaceholder{12.0000} & \stdplaceholder{0.1200} & \stdplaceholder{0.05} & \stdplaceholder{0.0100} & \stdplaceholder{6.0000} & \stdplaceholder{0.0700} & \stdplaceholder{0.02} & \stdplaceholder{0.0080} & \stdplaceholder{5.5000} & \stdplaceholder{0.0500} & \stdplaceholder{0.02} \\
    \rowcolor[gray]{0.92} CoRe \cite{yu2021group} & ICCV'21 & 0.8432 & \underline{85.5388} & \underline{0.8554} & 11.53 & 0.7520 & \textbf{124.9170} & \textbf{2.0470} & 3.41 & \underline{0.6545} & \textbf{365.8256} & \textbf{4.2427} & 3.47 & 0.5328 & 516.1882 & 5.1619 & 4.17 & 0.9061 & 108.2117 & 1.3305 & 2.02 & 0.8452 & 337.7187 & 3.5469 & 2.03 \\
    \rowcolor[gray]{0.98}  &  & \stdplaceholder{0.0145} & \stdplaceholder{14.6279} & \stdplaceholder{0.1463} & \stdplaceholder{0.01} & \stdplaceholder{0.0047} & \stdplaceholder{6.3346} & \stdplaceholder{0.1038} & \stdplaceholder{0.03} & \stdplaceholder{0.0263} & \stdplaceholder{20.1799} & \stdplaceholder{0.2340} & \stdplaceholder{0.02} & \stdplaceholder{0.0689} & \stdplaceholder{49.8321} & \stdplaceholder{0.4983} & \stdplaceholder{0.04} & \stdplaceholder{0.0147} & \stdplaceholder{17.8071} & \stdplaceholder{0.2189} & \stdplaceholder{0.02} & \stdplaceholder{0.0433} & \stdplaceholder{29.1758} & \stdplaceholder{0.3064} & \stdplaceholder{0.01} \\
    \rowcolor[gray]{0.92} GDLT \cite{xu2022likert} & CVPR'22 & 0.8425 & \textbf{82.4375} & \textbf{0.8244} & 3.77 & 0.7665 & 159.1891 & 2.6086 & 1.50 & 0.6540 & 397.9806 & 4.6157 & 1.56 & 0.5805 & 526.6038 & 5.2661 & 1.79 & 0.9078 & \underline{64.1058} & \underline{0.7882} & \underline{0.85} & 0.9049 & 207.0555 & 2.1746 & 0.86 \\
    \rowcolor[gray]{0.98}  &  & \stdplaceholder{0.0104} & \stdplaceholder{4.2804} & \stdplaceholder{0.0428} & \stdplaceholder{0.01} & \stdplaceholder{0.0137} & \stdplaceholder{14.8464} & \stdplaceholder{0.2433} & \stdplaceholder{0.01} & \stdplaceholder{0.0319} & \stdplaceholder{52.1782} & \stdplaceholder{0.6052} & \stdplaceholder{0.01} & \stdplaceholder{0.0433} & \stdplaceholder{88.8919} & \stdplaceholder{0.8889} & \stdplaceholder{0.01} & \stdplaceholder{0.0095} & \stdplaceholder{3.2000} & \stdplaceholder{0.0400} & \stdplaceholder{0.01} & \stdplaceholder{0.0110} & \stdplaceholder{8.5000} & \stdplaceholder{0.0900} & \stdplaceholder{0.02} \\
    \rowcolor[gray]{0.92} HGCN \cite{zhou2023hierarchical} & TCSVT'23 & \textbf{0.8871} & 101.4813 & 1.0148 & \textbf{3.16} & \textbf{0.7725} & 220.5605 & 3.6143 & \textbf{1.36} & \textbf{0.6701} & 483.7507 & 5.6104 & \textbf{1.40} & \underline{0.6487} & \underline{369.5956} & \underline{3.6960} & \textbf{1.59} & 0.9174 & 96.7002 & 1.1890 & 0.86 & \textbf{0.9507} & \underline{140.0894} & \underline{1.4713} & \textbf{0.85} \\
    \rowcolor[gray]{0.98}  &  & \stdplaceholder{0.0110} & \stdplaceholder{3.2000} & \stdplaceholder{0.0250} & \stdplaceholder{0.05} & \stdplaceholder{0.0120} & \stdplaceholder{8.9000} & \stdplaceholder{0.1200} & \stdplaceholder{0.05} & \stdplaceholder{0.0150} & \stdplaceholder{18.0000} & \stdplaceholder{0.1800} & \stdplaceholder{0.05} & \stdplaceholder{0.0180} & \stdplaceholder{13.0000} & \stdplaceholder{0.1300} & \stdplaceholder{0.05} & \stdplaceholder{0.0090} & \stdplaceholder{4.5000} & \stdplaceholder{0.0500} & \stdplaceholder{0.02} & \stdplaceholder{0.0075} & \stdplaceholder{5.8000} & \stdplaceholder{0.0600} & \stdplaceholder{0.02} \\
    \rowcolor[gray]{0.92} DAE \cite{zhang2024auto} & NCAA'24 & 0.8468 & 99.1924 & 0.9919 & 3.27 & 0.7526 & 158.1046 & 2.5908 & \underline{1.43} & 0.5439 & 558.6434 & 6.4789 & 1.55 & 0.4824 & 589.9491 & 5.8995 & 3.23 & \textbf{0.9329} & 160.5100 & 1.9735 & 1.27 & 0.8883 & 278.3217 & 2.9231 & 1.19 \\
    \rowcolor[gray]{0.98}  &  & \stdplaceholder{0.0038} & \stdplaceholder{13.7533} & \stdplaceholder{0.1376} & \stdplaceholder{0.02} & \stdplaceholder{0.0152} & \stdplaceholder{45.3578} & \stdplaceholder{0.7433} & \stdplaceholder{0.01} & \stdplaceholder{0.0139} & \stdplaceholder{45.1156} & \stdplaceholder{0.5232} & \stdplaceholder{0.02} & \stdplaceholder{0.0341} & \stdplaceholder{17.2916} & \stdplaceholder{0.1729} & \stdplaceholder{0.01} & \stdplaceholder{0.0060} & \stdplaceholder{55.4437} & \stdplaceholder{0.6817} & \stdplaceholder{0.01} & \stdplaceholder{0.0086} & \stdplaceholder{17.2860} & \stdplaceholder{0.1815} & \stdplaceholder{0.01} \\
    \rowcolor[gray]{0.92} T$^2$CR \cite{ke2024two} & INFS‘24 & 0.8334 & 96.7862 & 0.9679 & 9.13 & 0.7585 & 187.0455 & 3.0651 & 4.93 & 0.5887 & 482.7082 & 5.5983 & 5.03 & 0.4790 & 546.7196 & 5.4672 & 5.55 & 0.9087 & 242.4116 & 2.9805 & 2.58 & 0.9109 & 265.3465 & 2.7868 & 2.71 \\
    \rowcolor[gray]{0.98}  &  & \stdplaceholder{0.0140} & \stdplaceholder{4.1000} & \stdplaceholder{0.0300} & \stdplaceholder{0.10} & \stdplaceholder{0.0135} & \stdplaceholder{7.5000} & \stdplaceholder{0.1400} & \stdplaceholder{0.07} & \stdplaceholder{0.0160} & \stdplaceholder{17.5000} & \stdplaceholder{0.1900} & \stdplaceholder{0.08} & \stdplaceholder{0.0200} & \stdplaceholder{19.0000} & \stdplaceholder{0.2000} & \stdplaceholder{0.07} & \stdplaceholder{0.0105} & \stdplaceholder{9.5000} & \stdplaceholder{0.1100} & \stdplaceholder{0.05} & \stdplaceholder{0.0105} & \stdplaceholder{10.5000} & \stdplaceholder{0.1100} & \stdplaceholder{0.05} \\
    \rowcolor[gray]{0.92} CoFInAl \cite{zhou2024cofinal} & IJCAI'24 & 0.8652 & 150.5001 & 1.5050 & \underline{3.17} & \underline{0.7685} & \underline{126.4775} & \underline{2.0726} & 1.43 & 0.6119 & \underline{394.2921} & \underline{4.5729} & 1.45 & 0.5990 & 539.6546 & 5.3965 & 1.64 & 0.9073 & \textbf{46.2191} & \textbf{0.5683} & 0.85 & 0.9334 & 243.4674 & 2.5570 & 0.87 \\
    \rowcolor[gray]{0.98}  &  & \stdplaceholder{0.0130} & \stdplaceholder{7.1000} & \stdplaceholder{0.0550} & \stdplaceholder{0.06} & \stdplaceholder{0.0125} & \stdplaceholder{5.2000} & \stdplaceholder{0.0850} & \stdplaceholder{0.04} & \stdplaceholder{0.0170} & \stdplaceholder{14.0000} & \stdplaceholder{0.1600} & \stdplaceholder{0.05} & \stdplaceholder{0.0150} & \stdplaceholder{18.0000} & \stdplaceholder{0.1800} & \stdplaceholder{0.05} & \stdplaceholder{0.0100} & \stdplaceholder{2.2000} & \stdplaceholder{0.0300} & \stdplaceholder{0.02} & \stdplaceholder{0.0090} & \stdplaceholder{9.0000} & \stdplaceholder{0.0900} & \stdplaceholder{0.02} \\
    \midrule
    \multirow{2.5}{*}{\textbf{Method}} & \multirow{2.5}{*}{\textbf{Publisher}} & \multicolumn{4}{c}{\textbf{Ball}  $\color{gray}^\text{RG}$ \cite{zeng2020hybrid}} & \multicolumn{4}{c}{\textbf{Clubs} $\color{gray}^\text{RG}$ \cite{zeng2020hybrid}} & \multicolumn{4}{c}{\textbf{Hoop}  $\color{gray}^\text{RG}$ \cite{zeng2020hybrid}} & \multicolumn{4}{c}{\textbf{Ribbon} $\color{gray}^\text{RG}$ \cite{zeng2020hybrid}} & \multicolumn{4}{c}{\textbf{TES} $\color{gray}^\text{Fis-V}$ \cite{xu2019learning}} & \multicolumn{4}{c}{\textbf{PCS} $\color{gray}^\text{Fis-V}$ \cite{xu2019learning}} \\ 
    \cmidrule(lr){3-6} \cmidrule(lr){7-10} \cmidrule(lr){11-14} \cmidrule(lr){15-18} \cmidrule(lr){19-22} \cmidrule(lr){23-26}
    & & SRCC & MSE & rMSE & Time & SRCC & MSE & rMSE & Time & SRCC & MSE & rMSE & Time & SRCC & MSE & rMSE & Time & SRCC & MSE & rMSE & Time & SRCC & MSE & rMSE & Time \\
    \midrule
    \rowcolor[gray]{0.92} MUSDL \cite{tang2020uncertainty} & CVPR'20 & 0.4897 & 435.5650 & 4.3557 & \textbf{0.01} & 0.4746 & 439.7667 & 4.3977 & \textbf{0.01} & 0.4911 & 452.2267 & 4.5223 & \textbf{0.01} & 0.6128 & 466.8100 & 4.6681 & \textbf{0.01} & 0.0000 & 7823.4400 & 78.2344 & \textbf{0.02} & 0.6506 & 407.5300 & 4.0753 & \textbf{0.02} \\
    \rowcolor[gray]{0.98}  &  & \stdplaceholder{0.0503} & \stdplaceholder{1.1513} & \stdplaceholder{0.3075} & \stdplaceholder{0.00} & \stdplaceholder{0.0481} & \stdplaceholder{0.5349} & \stdplaceholder{0.2509} & \stdplaceholder{0.00} & \stdplaceholder{0.0230} & \stdplaceholder{1.4687} & \stdplaceholder{0.5528} & \stdplaceholder{0.00} & \stdplaceholder{0.0535} & \stdplaceholder{0.7110} & \stdplaceholder{0.2660} & \stdplaceholder{0.00} & \stdplaceholder{0.0000} & \stdplaceholder{0.0000} & \stdplaceholder{0.0000} & \stdplaceholder{0.00} & \stdplaceholder{0.0225} & \stdplaceholder{0.7409} & \stdplaceholder{0.1184} & \stdplaceholder{0.00} \\
    \rowcolor[gray]{0.92} CoRe \cite{yu2021group} & ICCV'21 & 0.7510 & 160.1400 & 1.6014 & 0.06 & \textbf{0.7592} & \underline{232.4600} & \textbf{2.3246} & 0.06 & 0.7554 & 211.4200 & 2.1148 & 0.12 & 0.7439 & 246.2800 & 2.4628 & 0.06 & 0.6578 & 186.3000 & \underline{1.8630} & 0.20 & 0.7812 & 181.6600 & 1.8166 & 0.20 \\
    \rowcolor[gray]{0.98}  &  & \stdplaceholder{0.0323} & \stdplaceholder{0.7252} & \stdplaceholder{0.1937} & \stdplaceholder{0.02} & \stdplaceholder{0.0260} & \stdplaceholder{0.4628} & \stdplaceholder{0.2171} & \stdplaceholder{0.00} & \stdplaceholder{0.0049} & \stdplaceholder{0.2195} & \stdplaceholder{0.0827} & \stdplaceholder{0.00} & \stdplaceholder{0.0174} & \stdplaceholder{0.2478} & \stdplaceholder{0.0927} & \stdplaceholder{0.00} & \stdplaceholder{0.0099} & \stdplaceholder{0.4128} & \stdplaceholder{0.0347} & \stdplaceholder{0.01} & \stdplaceholder{0.0065} & \stdplaceholder{0.7223} & \stdplaceholder{0.1154} & \stdplaceholder{0.02} \\
    \rowcolor[gray]{0.92} GDLT \cite{xu2022likert} & CVPR'22 & \textbf{0.7759} & 197.7200 & 1.9772 & \underline{0.02} & 0.7365 & 263.7000 & 2.6370 & \underline{0.02} & \underline{0.7616} & 179.1000 & \underline{1.7910} & 0.04 & \textbf{0.8007} & \underline{202.0100} & \textbf{2.0201} & \underline{0.02} & 0.6304 & 237.0700 & 2.3707 & 0.04 & 0.7911 & \underline{143.3400} & \textbf{1.4334} & 0.04 \\
    \rowcolor[gray]{0.98}  &  & \stdplaceholder{0.0255} & \stdplaceholder{0.5975} & \stdplaceholder{0.1595} & \stdplaceholder{0.00} & \stdplaceholder{0.0085} & \stdplaceholder{0.3780} & \stdplaceholder{0.1774} & \stdplaceholder{0.00} & \stdplaceholder{0.0278} & \stdplaceholder{0.6003} & \stdplaceholder{0.2259} & \stdplaceholder{0.00} & \stdplaceholder{0.0034} & \stdplaceholder{0.3356} & \stdplaceholder{0.1255} & \stdplaceholder{0.00} & \stdplaceholder{0.0202} & \stdplaceholder{0.7249} & \stdplaceholder{0.0608} & \stdplaceholder{0.00} & \stdplaceholder{0.0163} & \stdplaceholder{1.2498} & \stdplaceholder{0.1996} & \stdplaceholder{0.00} \\
    \rowcolor[gray]{0.92} HGCN \cite{zhou2023hierarchical} & TCSVT'23 & 0.7007 & 197.3500 & 1.9735 & 0.02 & \underline{0.7585} & 246.6900 & 2.4669 & 0.02 & \textbf{0.7701} & \underline{176.4000} & \textbf{1.7640} & \underline{0.03} & 0.7603 & 215.9700 & \underline{2.1597} & 0.02 & \underline{0.6830} & 273.5100 & 2.7351 & \underline{0.03} & 0.7657 & 182.2100 & 1.8221 & \underline{0.03} \\
    \rowcolor[gray]{0.98}  &  & \stdplaceholder{0.0135} & \stdplaceholder{0.3744} & \stdplaceholder{0.1000} & \stdplaceholder{0.00} & \stdplaceholder{0.0550} & \stdplaceholder{0.8109} & \stdplaceholder{0.3804} & \stdplaceholder{0.00} & \stdplaceholder{0.0352} & \stdplaceholder{1.3832} & \stdplaceholder{0.5206} & \stdplaceholder{0.00} & \stdplaceholder{0.0532} & \stdplaceholder{2.3762} & \stdplaceholder{0.8889} & \stdplaceholder{0.00} & \stdplaceholder{0.0179} & \stdplaceholder{5.2356} & \stdplaceholder{0.4394} & \stdplaceholder{0.00} & \stdplaceholder{0.0101} & \stdplaceholder{0.2847} & \stdplaceholder{0.0454} & \stdplaceholder{0.00} \\
    \rowcolor[gray]{0.92} DAE \cite{zhang2024auto} & NCAA'24 & 0.7425 & \underline{143.1000} & \textbf{1.4310} & 0.02 & 0.7518 & 402.1000 & 4.0210 & 0.02 & 0.7431 & 183.4900 & 1.8349 & 0.03 & 0.7399 & 388.5100 & 3.8851 & 0.02 & 0.6748 & 198.8100 & 1.9881 & 0.03 & \underline{0.8014} & 163.5100 & \underline{1.6351} & 0.03 \\
    \rowcolor[gray]{0.98}  &  & \stdplaceholder{0.0037} & \stdplaceholder{0.4228} & \stdplaceholder{0.1129} & \stdplaceholder{0.00} & \stdplaceholder{0.0040} & \stdplaceholder{1.0763} & \stdplaceholder{0.5049} & \stdplaceholder{0.00} & \stdplaceholder{0.0022} & \stdplaceholder{0.1391} & \stdplaceholder{0.0523} & \stdplaceholder{0.00} & \stdplaceholder{0.0030} & \stdplaceholder{0.6537} & \stdplaceholder{0.2445} & \stdplaceholder{0.00} & \stdplaceholder{0.0058} & \stdplaceholder{1.9698} & \stdplaceholder{0.1653} & \stdplaceholder{0.00} & \stdplaceholder{0.0058} & \stdplaceholder{2.2894} & \stdplaceholder{0.3657} & \stdplaceholder{0.00} \\
    \rowcolor[gray]{0.92} T$^2$CR \cite{ke2024two} & INFS‘24 & 0.6483 & 185.0900 & 1.8509 & 0.28 & 0.7335 & 237.6300 & \underline{2.3763} & 0.28 & 0.6810 & 284.4400 & 2.8444 & 0.28 & \underline{0.7769} & 231.5000 & 2.3150 & 0.28 & \textbf{0.6896} & \underline{184.9400} & \textbf{1.8494} & 1.14 & 0.7536 & 178.8300 & 1.7883 & 1.05 \\
    \rowcolor[gray]{0.98}  &  & \stdplaceholder{0.0178} & \stdplaceholder{0.3949} & \stdplaceholder{0.1055} & \stdplaceholder{0.00} & \stdplaceholder{0.0065} & \stdplaceholder{0.3280} & \stdplaceholder{0.1539} & \stdplaceholder{0.05} & \stdplaceholder{0.0133} & \stdplaceholder{0.6959} & \stdplaceholder{0.2619} & \stdplaceholder{0.01} & \stdplaceholder{0.0314} & \stdplaceholder{0.3198} & \stdplaceholder{0.1196} & \stdplaceholder{0.02} & \stdplaceholder{0.0132} & \stdplaceholder{0.8095} & \stdplaceholder{0.0679} & \stdplaceholder{0.09} & \stdplaceholder{0.0130} & \stdplaceholder{0.6193} & \stdplaceholder{0.0989} & \stdplaceholder{0.03} \\
    \rowcolor[gray]{0.92} CoFInAl \cite{zhou2024cofinal} & IJCAI'24 & \underline{0.7616} & \textbf{5.9593} & \underline{1.5916} & 0.14 & 0.7443 & \textbf{6.4109} & 3.0076 & 0.12 & 0.7482 & \textbf{24.4233} & 9.1924 & 0.11 & 0.7593 & \textbf{6.4777} & 2.4232 & 0.06 & 0.6539 & \textbf{37.2062} & 3.1223 & 0.06 & \textbf{0.8152} & \textbf{17.4047} & 2.7803 & 0.12 \\
    \rowcolor[gray]{0.98}  &  & \stdplaceholder{0.0220} & \stdplaceholder{1.4496} & \stdplaceholder{0.3872} & \stdplaceholder{0.00} & \stdplaceholder{0.0113} & \stdplaceholder{0.6054} & \stdplaceholder{0.2840} & \stdplaceholder{0.02} & \stdplaceholder{0.0144} & \stdplaceholder{2.0579} & \stdplaceholder{0.7746} & \stdplaceholder{0.00} & \stdplaceholder{0.0039} & \stdplaceholder{0.2496} & \stdplaceholder{0.0934} & \stdplaceholder{0.00} & \stdplaceholder{0.0102} & \stdplaceholder{8.1239} & \stdplaceholder{0.6818} & \stdplaceholder{0.00} & \stdplaceholder{0.0199} & \stdplaceholder{2.9021} & \stdplaceholder{0.4636} & \stdplaceholder{0.01} \\
    \bottomrule
    \end{tabular}
    }
    \label{tab:benchmark_results}
\end{table*}

\myPara{Evaluation Metrics}
We report SRCC for ranking, MSE, and rMSE for accuracy, as well as computational metrics such as training time, model size, inference speed, and computational complexity, providing a balanced view of accuracy and efficiency.

\myPara{Implementation Details}
All benchmark results are \textbf{reproduced} using publicly available implementations under a unified evaluation protocol.
Unless otherwise specified, experiments are conducted on a single RTX~3090 GPU with an I3D backbone pre-trained on Kinetics-400, trained for 100 epochs using the Adam optimizer (learning rate $1\times10^{-4}$, weight decay $1\times10^{-4}$). For datasets without released configurations, we follow official settings from prior works; for RG and Fis-V, official VST features are directly adopted. To ensure fairness and reproducibility, no per-dataset or per-method hyperparameter tuning is performed. Results are averaged over three runs (seeds 1022, 1023, and 1024) on the official splits, following common AQA benchmarking practice. Task-specific tuning may further improve individual methods.
 
\subsubsection{Assessment Performance}
\cref{tab:benchmark_results} compares seven AQA methods across six benchmark datasets. Results are based on standardized implementations with official code for fair comparison, and the reported standard deviations indicate generally stable performance across runs.
 Transformer-based methods, such as CoFInAl~\cite{zhou2024cofinal}, perform competitively on several long-sequence datasets (e.g., RG, LOGO), while GDLT~\cite{xu2022likert} achieves strong SRCC performance on RG and LOGO. However, USDL/MUSDL~\cite{tang2020uncertainty} shows larger variance on challenging datasets such as Fis-V, suggesting sensitivity to noise and complex temporal dependencies.
Contrastive regression methods (CoRe~\cite{yu2021group}, T$^2$CR~\cite{ke2024two}) perform competitively on large-scale datasets (MTL-AQA, FineDiving), while their performance varies more across smaller datasets. Uncertainty-aware DAE~\cite{zhang2024auto} shows strong results on Fis-V, indicating the benefit of modeling prediction uncertainty.
GCN-based HGCN~\cite{zhou2023hierarchical} also performs competitively on AQA-7, suggesting the effectiveness of graph-based local and global feature modeling for robust performance. 
\begin{figure}[!t]
    \centering
    \sf\tiny
    \begin{minipage}[c]{\linewidth}
    \centering
    \begin{overpic}[width=\linewidth]{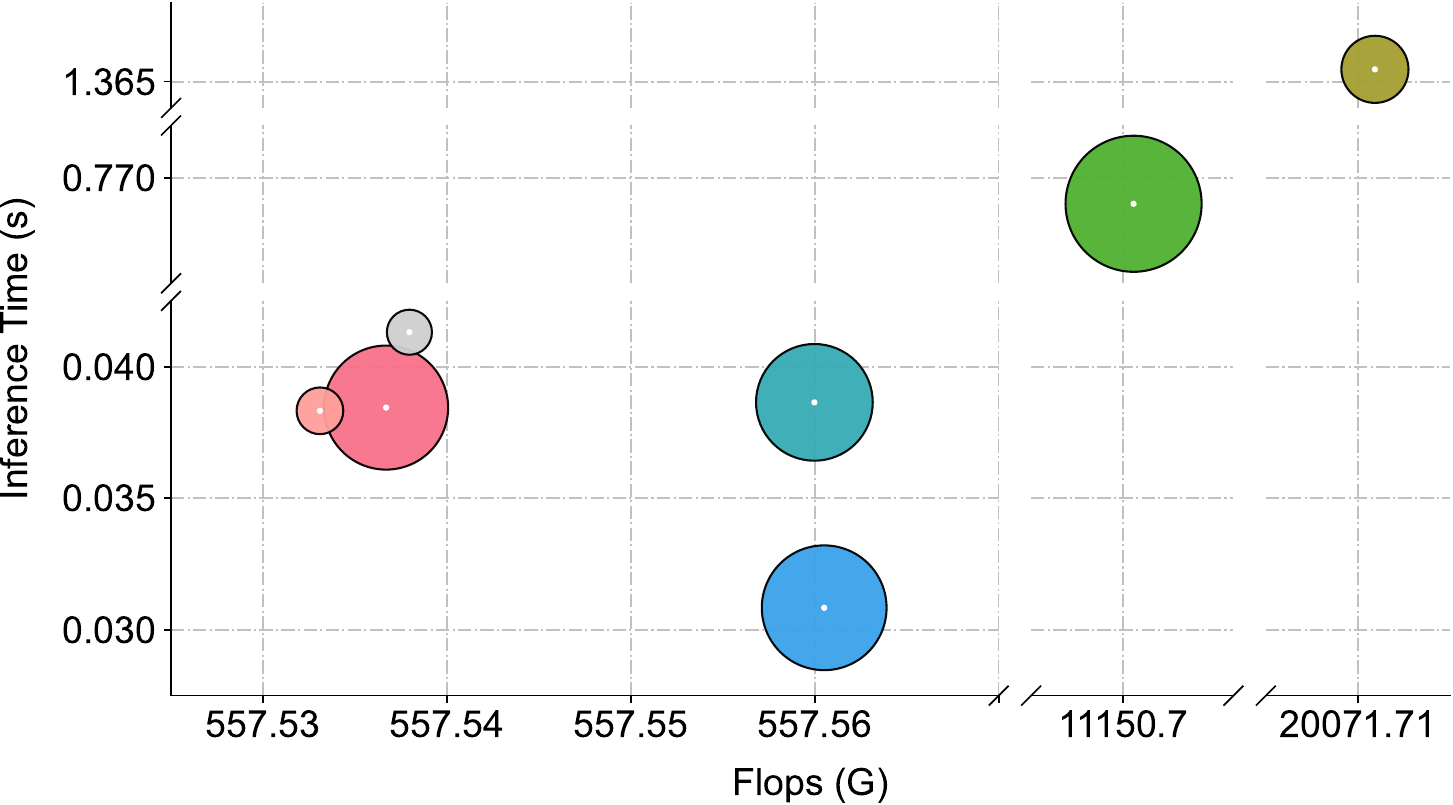}
        \put(24,36){  HGCN \cite{zhou2023hierarchical}}
        \put(59,40){  CoRe \cite{yu2021group}}
        \put(78,50){  T$^2$CR \cite{ke2024two}}
        \put(62,27){  GDLT \cite{xu2022likert}}
        \put(62,12){  CoFInAl \cite{zhou2024cofinal}}
        \put(32,26){  MUSDL \cite{tang2020uncertainty}}
        \put(13,22){  DAE \cite{zhang2024auto}}
        \put(17,49.5){  GCN-Based}
        \put(25,47){\linethickness{0.10mm}\color{gray}\vector(1,-2.8){3}}
        \put(16,48){\tikz\fill[gray,fill opacity=0.1,rounded corners](0,0)rectangle(1.20,0.35);}
        \put(16,48){\tikz\draw[thick,color=white,line width=0.00mm,rounded corners] (0,0)rectangle(1.20,0.35);}
        \put(39,48){\tikz\fill[gray,fill opacity=0.1,rounded corners](0,0)rectangle(2.0,0.35);}
        \put(39,48){\tikz\draw[thick,color=white,line width=0.00mm,rounded corners] (0,0)rectangle(2.0,0.35);}
        \put(40,49.5){  Contrastive Regression}
        \put(55,47){\linethickness{0.10mm}\color{gray}\vector(1,-1.2){4}}
        \put(70.8,51){\linethickness{0.10mm}\color{gray}\vector(1,0){7}}
        \put(75.5,20){  Transformer-Based}
        \put(74.8,18){\tikz\fill[gray,fill opacity=0.1,rounded corners](0,0)rectangle(1.70,0.35);}
        \put(74.8,18){\tikz\draw[thick,color=white,line width=0.00mm,rounded corners] (0,0)rectangle(1.70,0.35);}
        \put(74,20){\linethickness{0.10mm}\color{gray}\vector(-1,-0.8){7}}
        \put(74,21){\linethickness{0.10mm}\color{gray}\vector(-1, 0.8){7}}
        \put(20,13){  Uncertainty-Aware}
        \put(19,11.3){\tikz\fill[gray,fill opacity=0.1,rounded corners](0,0)rectangle(1.70,0.35);}
        \put(19,11.3){\tikz\draw[thick,color=white,line width=0.00mm,rounded corners] (0,0)rectangle(1.70,0.35);}
        \put(29,16.6){\linethickness{0.10mm}\color{gray}\vector(-1,0.5){8}}
        \put(32,16.6){\linethickness{0.10mm}\color{gray}\vector(1,2.8){3}}
    \end{overpic}
    \end{minipage}\hfill
    \begin{minipage}[c]{\linewidth}
        \caption{Computation comparison with selected baselines under the identical device settings on the MTL-AQA dataset. The x-axis represents the FLOPs, the y-axis indicates the inference time, and the bubble size corresponds to the number of parameters.}
        \label{fig:bubble-plot}
    \end{minipage}
\end{figure}

\subsubsection{Computational Performance}
\cref{tab:benchmark_results} and \cref{fig:bubble-plot} compare training time, FLOPs, parameters, and inference speed across methods (statistics from MTL-AQA). 
Contrastive regression methods (CoRe~\cite{yu2021group}, T$^2$CR~\cite{ke2024two}) are the most computationally intensive, with training times of 46 and 49 hours, respectively, and the longest inference times (up to 1.4s per sample), mainly due to exemplar computation and ensemble strategies. T$^2$CR’s dual time-scale design further increases computation. In contrast, direct regression and transformer-based methods (e.g., CoFInAl~\cite{zhou2024cofinal}) are more efficient, with CoFInAl showing the fastest inference (0.03s). HGCN~\cite{zhou2023hierarchical} and DAE~\cite{zhang2024auto} have the lowest parameter counts (around 12.5M), and DAE has the lowest FLOPs (557.53G), whereas T$^2$CR’s complexity is over 30 times higher.

\subsubsection{Experimental Observations}
As shown in \cref{tab:benchmark_results,fig:bubble-plot}, our benchmark provides several key insights into the strengths and limitations of existing AQA methods. First, no single method consistently achieves the best performance across all datasets; methods that perform well on one dataset often fail to maintain the same advantage on others. Second, performance is strongly influenced by dataset characteristics, particularly differences in temporal length and domain structure. Third, higher predictive accuracy is often accompanied by increased computational cost, indicating a clear trade-off between performance and efficiency. These findings highlight fundamental limitations of existing AQA approaches and emphasize the need for models that are both computationally efficient and robust across diverse datasets.

\begin{figure*}[!h]
    \centering
    \begin{minipage}[c]{\linewidth}
        \includegraphics[width=\linewidth,clip=10 10 10 10]{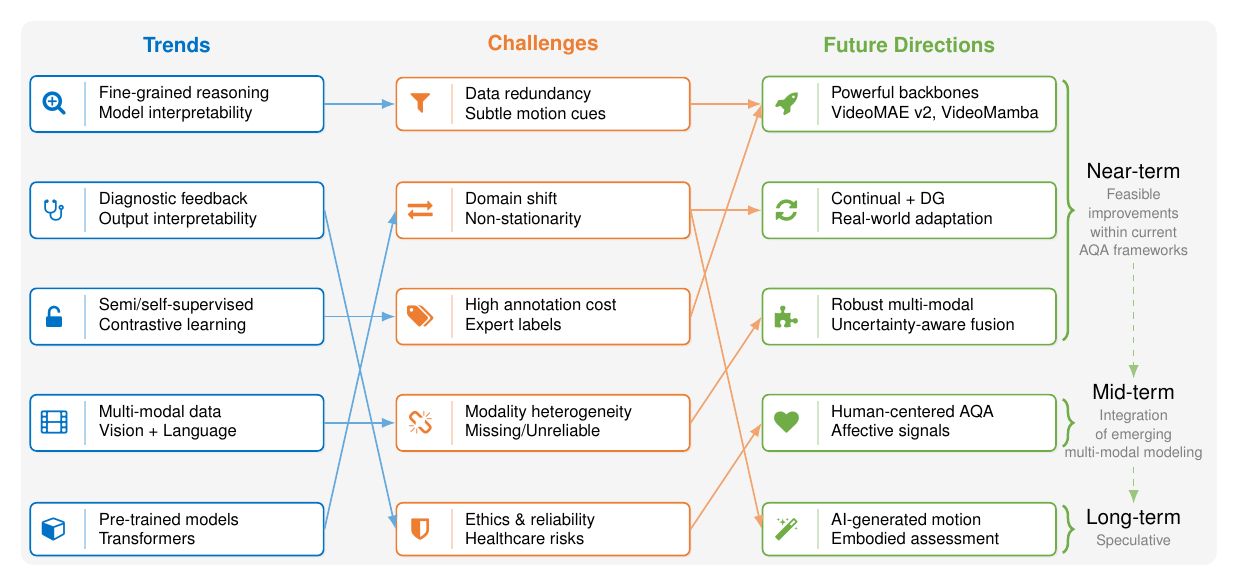}
    \end{minipage}
    \hfill
    \begin{minipage}[c]{\linewidth}
         \caption{Evolutionary roadmap of AQA. The diagram connects current methodological \textbf{trends} (left) with emerging \textbf{challenges} (middle), and highlights potential \textbf{future directions} (right).}
         \label{fig:aqa_roadmap}
\end{minipage}
    \vspace{0cm}
\end{figure*}

\section{Trends, Challenges, and Future Directions}
\label{sec:discussion}

As shown in \cref{fig:aqa_roadmap}, AQA is evolving beyond benchmark-driven accuracy toward more robust and adaptive systems for real-world deployment. This section synthesizes key methodological trends, remaining challenges, and future directions in AQA. 

\subsection{Current Trends}
\label{sec:trends}
Recent AQA research is shifting from simple score regression toward structure-aware modeling. 
A prominent trend is the adoption of \textbf{fine-grained reasoning}~\cite{xu2022finediving,xu2024fineparser,han2025finecausal} to support model-level interpretability, where internal representations are aligned with meaningful temporal segments or sub-actions. 
In parallel, increasing attention is paid to \textbf{output-level interpretability}~\cite{ashutosh2024expertaf,li2024techcoach}, moving beyond scalar scores toward diagnostic and actionable feedback. 
Driven by limited annotations, AQA is increasingly integrated with \textbf{semi/self-supervised} learning \cite{yun2024semi,gedamu2024self} to better exploit sparse labels, \textbf{contrastive} objectives \cite{yu2021group,xu2024fineparser} to enhance assessment robustness, and \textbf{multi-modal} data \cite{xu2024vision,xu2025language} to leverage complementary cues. 
At the architectural level, transformers \cite{bai2022action,xu2022likert} and \textbf{pre-trained models} \cite{carreira2017quo,liu2022video} are becoming common backbones, enabling long-range temporal modeling and improved data efficiency. 
Together, these trends indicate a transition toward more interpretable and generalizable AQA systems.

\subsection{Under-explored Challenges}
\label{sec:chellenges}

Despite recent advances, several challenges in AQA remain under-explored.

\myPara{Data Redundancy versus Subtle Cues}  
High-dimensional video inputs contain considerable redundancy, while action quality depends on subtle motion variations. Balancing information suppression and cue amplification remains a core challenge.

\myPara{Domain Shift and Non-stationarity}  
Gaps between pre-trained domains and AQA tasks, together with real-world non-stationary variations, significantly degrade model performance. Effectively adapting to such shifts remains largely unsolved.

\myPara{High Annotation Cost}  
Fine-grained quality annotations, such as stage-level or temporal labels, require expert knowledge and are expensive to obtain. Learning reliable quality representations from coarse or sparse supervision remains challenging.

\myPara{Modality Heterogeneity}  
Although multi-modal AQA improves performance, real-world data often involve missing or unreliable modalities. Designing robust models that gracefully degrade under incomplete multi-modal inputs remains under-explored.

\myPara{Ethics and Reliability}  
In sensitive domains such as healthcare and rehabilitation, AQA systems must ensure reliability, fairness, and interpretability. Addressing ethical risks and enabling responsible deployment remain open challenges.

\subsection{Future Directions}
\label{sec:future}
Building on the above challenges, we outline several promising directions for advancing AQA, organized roughly from near-term extensions of existing pipelines to longer-term opportunities enabled by emerging AI paradigms (see \cref{fig:aqa_roadmap}).

\myPara{Integrating Powerful Foundation Models} 
Our analysis in \cref{sec:taxonomy} reveals that progress in AQA closely follows backbone evolution (see \cref{fig:backbone_evolution_2012_2025}), and recent works~\cite{zhou2024cofinal,zhou2025phi} show that stronger backbones consistently improve performance. 
Meanwhile, our benchmark analysis shows that current AQA datasets are limited by label scarcity (see \cref{sec:db}), which constrains supervised learning and limits generalization. This motivates the adoption of foundation models that leverage large-scale pre-training to learn transferable representations,  such as VideoMAE V2~\cite{wang2023videomae} and VideoMamba~\cite{park2024videomamba} for spatiotemporal representation, MotionGPT~\cite{jiang2023motiongpt} for skeletal modeling, and Video-LLaMA~\cite{zhang2023video} for video-language understanding. While these models provide richer representations and cross-modal reasoning capabilities, effective adaptation to AQA-specific requirements remains an open challenge.

\myPara{Continual Learning and Domain Generalization}  
Existing continual AQA studies~\cite{li2024continual,zhou2025continual,zhou2026brima} still largely rely on labeled data for model updates. 
Future research should jointly address continual adaptation and generalization to unseen domains, particularly in ego-view settings~\cite{li2024egoexo}, where viewpoints and motion patterns evolve significantly over time, enabling robust AQA under non-stationary conditions. 
Moreover, these methods rely on rehearsal strategies that store previously seen data, which may raise privacy concerns and limit their applicability in real-world scenarios.

\myPara{Robust Multi-modal Learning}  
Our analysis in \cref{sec:taxonomy} reveals that most multi-modal AQA methods assume complete and reliable modality inputs, limiting their applicability in real-world scenarios where modalities may be missing or noisy. This motivates the design of AQA models that are robust to incomplete or unreliable modalities \cite{xu2025mcmoe,zhou2026brima}. Future work should focus on flexible architectures with uncertainty-aware fusion and graceful degradation under partial multi-modal inputs.

\myPara{Human-centered and Trustworthy AQA}  
As discussed in \cref{sec:app} and \cref{sec:db}, AQA is often applied in sensitive domains such as healthcare \cite{zhou2023video}, where reliability and interpretability are critical. However, current methods remain vulnerable to adversarial perturbations and lack sufficient interpretability for practical deployment.  This motivates the development of trustworthy AQA systems that emphasize robustness, interpretability, and ethical considerations. Future work should also incorporate affective signals, such as emotion and engagement, to enable more human-centered assessment.

\myPara{AI-generated and Embodied Action Assessment}  
The rapid progress of action generation foundation models~\cite{jiang2023motiongpt,yuan2023physdiff} has enabled realistic motion synthesis, making reliable evaluation increasingly important. 
In this context, AQA can play a critical role in assessing the realism, correctness, and execution quality of generated motions for applications such as e-commerce avatars and AI hosts~\cite{chen2024gaia}. 
Beyond motion synthesis, similar evaluation challenges arise in embodied learning and robotics, where agents must perform actions in physical or simulated environments~\cite{paolo2024position,feng2025multi}. 
Related problems have also been studied in surgical skill assessment, where methods are used to evaluate how humans operate robotic or medical systems~\cite{gao2023automatic,laughlin2025technical,boal2024evaluation}. 
Future research should explore how AQA methods can be integrated with generative and embodied models to provide automatic and fine-grained evaluation, potentially forming a feedback loop that further improves motion generation and embodied action learning.

\section{Conclusion}
\label{sec:conclusion}

This survey reviews recent advances in AQA, proposes a modality-driven hierarchical taxonomy, and introduces a unified benchmark for evaluating both accuracy and computational efficiency. By organizing fragmented literature and inconsistent comparisons, we clarify the methodological evolution across modalities and applications. We also summarize key trends, open challenges, and future research directions.
This survey has several limitations. Some emerging application-specific datasets are not discussed in depth due to space constraints. In addition, the benchmark focuses on video-based AQA, as skeleton-based and multi-modal methods still lack standardized datasets and reproducible pipelines. Thus, the evaluation emphasizes fair comparison under official settings rather than extensive task-specific tuning.
We hope this survey serves as a useful reference and promotes more robust and unified AQA research.

\vspace{0.2cm}
\section*{Acknowledgments}
This work was supported in part by NSFC under Grant 62272019.

\bibliographystyle{elsarticle-num} 
\bibliography{refs-full}





\end{document}